\documentclass{article}

    \PassOptionsToPackage{numbers}{natbib}
    \usepackage[preprint]{neurips_data_2024}
\usepackage{microtype}
\usepackage{graphicx}
\usepackage{tabularx}
\usepackage{booktabs} % for professional tables

\usepackage{hyperref}

\newcommand{\loose}{\looseness=-1}
\usepackage{amsthm}
\usepackage{algorithm,algorithmic}
\usepackage{amsmath}
\usepackage{amssymb}
\usepackage{mathtools}
\usepackage{enumitem}
\setlist[itemize]{noitemsep, topsep=0pt, leftmargin=11pt}
\setlist[enumerate]{noitemsep, topsep=0pt, leftmargin=11pt}
\usepackage{wrapfig}
\usepackage{xcolor}
\usepackage{subcaption}
\usepackage{makecell}
\usepackage{wrapfig}
\usepackage{multirow}
\usepackage[capitalize,noabbrev]{cleveref}
\usepackage{soul}
\usepackage[textsize=tiny]{todonotes}
\usepackage{wrapfig}
\renewenvironment{quote}
  {\small\list{}{\rightmargin=.5cm \leftmargin=.5cm}%
   \item\relax}
  {\endlist}

\usepackage[capitalize,noabbrev]{cleveref}

\theoremstyle{plain}

\theoremstyle{definition}

\theoremstyle{remark}

\def\1{\mathbf{1}}

\DeclareMathOperator*{\argmax}{arg\,max}

\title{Humor in AI: Massive Scale Crowd-Sourced Preferences and Benchmarks for Cartoon Captioning}

\author{%
  Jifan Zhang$^{1}$\thanks{Equal contribution.}\,\,, 
  Lalit Jain$^{2*}$,
  Yang Guo$^{1*}$,
  Jiayi Chen$^{1}$\thanks{Equal contribution.}\,\,,
  Kuan Lok Zhou$^{1\dagger}$, \\
  \textbf{Siddharth Suresh$^1$, Andrew Wagenmaker$^2$, Scott Sievert$^1$, Timothy Rogers$^1$,}\\
  \textbf{Kevin Jamieson$^2$, Robert Mankoff$^3$, Robert Nowak$^1$}\\
  $^1$University of Wisconsin-Madison,
  $^2$University of Washington, Seattle,\\
  $^3$Air Mail and Cartoon Collections
  \\
  \texttt{lalitj@uw.edu, \{jifan,yguo\}@cs.wisc.edu}
}

\begin{document}

\maketitle

\begin{abstract}
We present a novel multimodal preference dataset for creative tasks, consisting of over 250 million human ratings on more than 2.2 million captions, collected through crowdsourcing rating data for The New Yorker's weekly cartoon caption contest over the past eight years. This unique dataset supports the development and evaluation of multimodal large language models and preference-based fine-tuning algorithms for humorous caption generation. We propose novel benchmarks for judging the quality of model-generated captions, utilizing both GPT4 and human judgments to establish ranking-based evaluation strategies. Our experimental results highlight the limitations of current fine-tuning methods, such as RLHF and DPO, when applied to creative tasks. Furthermore, we demonstrate that even state-of-the-art models like GPT4 and Claude currently underperform top human contestants in generating humorous captions. As we conclude this extensive data collection effort, we release the entire preference dataset to the research community, fostering further advancements in AI humor generation and evaluation.\loose
\end{abstract}

\section{Introduction}\label{sec:intro}

This paper presents a dataset and benchmark for investigating alignment in Large Language Models (LLMs). Our dataset contains over a quarter of a billion human ratings from the New Yorker's cartoon caption contest. Writing funny captions presents significant challenges due to the subjectivity of humor and variability in human judgments. This benchmark offers a unique challenge for AI alignment, reflecting complexities found in tasks where expert humans consistently outperform current AI systems. Our study examines fundamental questions about aligning them to generate funny captions similar to the winning captions that are most highly rated by the New Yorker readers.

We explore humor expression in LLMs, investigating whether these models can recognize humor and generate amusing captions that resonate with human audiences. While LLMs are not specifically designed for humor, their training on diverse content suggests a potential for humor recognition and expression. We propose a benchmark for evaluating a model's humor capabilities using advanced systems like GPT-4.

Our empirical analysis shows that current LLMs can generate humorous captions but significantly underperform compared to high-ranking human submissions in the New Yorker's caption contests. Generating successful captions requires multiple advanced capabilities: understanding of cultural references, recognition of humor patterns, logical reasoning, systematic planning, and visual analysis. The multi-component nature of caption generation makes this benchmark an effective test of broad LLM capabilities. Progress in aligning LLMs for this task will require both advancing these individual capabilities and developing methods to integrate them effectively. This benchmark therefore provides a comprehensive integration test for LLM capabilities.

Our main contributions and findings are:

\textbf{Dataset:} We present a large-scale dataset of human-rated cartoon captions from The New Yorker's weekly contest. Each week, the New Yorker hosts a contest with a new cartoon, where thousands submit their funny captions. Hundreds of thousands of ratings are collected for each contest, and the winning captions are determined by those receiving the highest ratings. This dataset enables researchers to explore humor generation in LLMs and represents the first large-scale dataset with human judgments for evaluating creative tasks. With over 250 million ratings, it offers diverse examples for studying humor expression and perception in AI systems.

\textbf{Benchmark:} We introduce new metrics for evaluating humor quality in LLM-generated content, using GPT4 and group based techniques. These metrics provide a standardized framework for assessing AI-generated humor. Our benchmark allows for systematic comparisons between human and AI-generated humor.

\textbf{Evaluation of State-of-the-Art Models:} We assess the performance of models such as GPT-4 and Claude in generating humorous content, comparing their outputs to human-generated examples. This analysis offers insights into the current capabilities and limitations of LLMs in humor generation, identifying areas of strength and potential improvement.

\textbf{Alignment Strategy Analysis:} We use our benchmark to evaluate various alignment strategies, including Reinforcement Learning from Human Feedback (RLHF), Direct Preference Optimization (DPO), and Best-of-N sampling (BoN). By comparing these strategies, we provide insights into their effectiveness in enhancing humor generation in LLMs and aligning AI systems with human preferences.

In summary, our paper advances LLM capabilities in humor evaluation and generation through a comprehensive dataset, a new evaluation benchmark, and analysis of model performance and alignment strategies. This work enhances our understanding of humor in AI systems and provides a foundation for future research in this field. We open-source our dataset and code as detailed in Appendix~\ref{apx:links}.

\begin{figure}
    \centering
    \includegraphics[width=\linewidth]{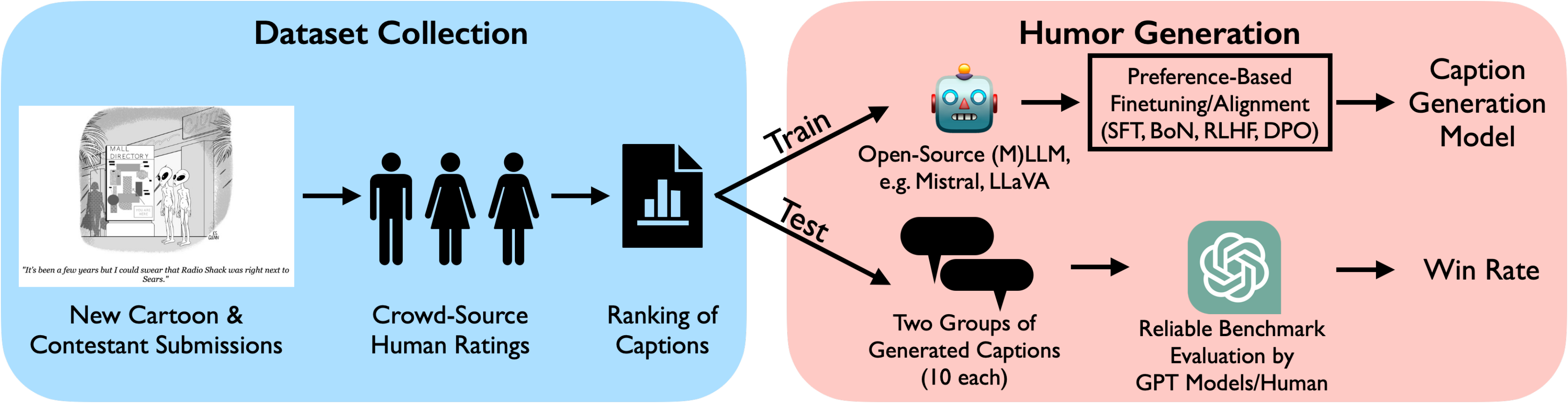}
    \caption{Overview of our workflow. During data collection, a new cartoon is released each week and thousands of captions are submitted. We then collect caption ratings through a crowd-sourcing procedure driven by a bandit algorithm. Our dataset is a collection of 365 contests, over 2.2M captions and over 250M human ratings. This dataset is utilized for our Humor generation task and benchmark. We experiment with finetuned open-source models and close-sourced API calls (both LLMs and MLLMs). Our novel and low-cost evaluator provides better reliability in evaluating captions.}
    \label{fig:overview}
\end{figure}

\section{Related Work}\label{sec:related_work}
\textbf{New Yorker Caption Contest.} Since its original conception as part of the NEXT crowdsourcing system~\cite{jamieson2015next, sievert2017next}, the New Yorker Caption Contest Dataset has been updated on a weekly basis for the last several years. During this time, the dataset has been primarily used for the evaluation of online algorithms and, similar to this work, to study the nature of humor. Works in the former camp include \cite{mason2020finding,tanczos2017kl,yang2017framework}. Perhaps the most relevant work to ours is ~\cite{hessel2022androids}. They formulated three tasks, matching, quality ranking and explanation generation for studying whether current AI systems \emph{understand} humor. 
%Besides studies on the dataset we introduced, prior works include 
Additional prior work includes ~\cite{shahaf2015inside,radev2015humor,king2013random}, which utilize judgements made by the editors of the New Yorker directly to analyze a smaller number of contests ($<50$) and attempt to identify features that correlate with caption performance such as length, perplexity, readability and sentiment. 
%These works utilize judgements made by the editors of the New Yorker directly. 

\textbf{Alignment of LLMs.}
Finetuning of LLMs has proved a critical step in aligning the behavior of pretrained models to downstream tasks. 
A standard pipeline  is to first finetune the pretrained model via \emph{supervised fine-tuning} (SFT)---to imitate expert demonstrations---followed by 
\emph{reinforcement learning from human feedback} (RLHF) \citep{christiano2017deep}---where a reward model is trained on human preferences, and then the SFT model is trained to maximize this reward via PPO \citep{schulman2017proximal}.
This pipeline has been successfully applied for finetuning frontier models \citep{ziegler2019fine,bai2022training,ouyang2022training,touvron2023llama}, and has inspired a vast amount of follow-up work refining and extending the SFT \citep{yuan2023rrhf,zhao2023slic,gulcehre2023reinforced,mukobi2023superhf,ethayarajh2024kto} and RLHF \citep{bakker2022fine,dumoulin2023density,liu2023statistical,siththaranjan2023distributional,munos2023nash,swamy2024minimaximalist,chakraborty2024maxmin,chang2024dataset} methodologies.
%While powerful, RLHF is typically very sensitive to hyperparameters and can be difficult to effectively apply. 
\emph{Direct preference optimization} (DPO) methods \citep{rafailov2024direct} have recently emerged as a simpler yet still effective replacement to the RLHF paradigm.
Instead of training a reward model and then optimizing this reward, DPO combines these steps by directly optimizing the SFT model on offline human preference data,
%In many cases this yields similar performance to RLHF, 
and has inspired a variety of extensions \citep{hejna2023contrastive,azar2024general,song2024preference,rosset2024direct,tang2024generalized,yin2024relative}.
While the aforementioned works focus on finetuning on human feedback, a related line of works has sought to finetune on AI-generated feedback \citep{yang2023rlcd,xu2023contrastive,lee2023rlaif,burns2023weak,chen2024self,yuan2024self,bai2022constitutional}.
Despite extensive research into various fine-tuning methodologies, understanding their effectiveness for creative tasks remains nascent. 
While several studies have explored when and why different methods are most effective \citep{gao2023scaling,kirk2023understanding,wu2024fine,chan2024dense,zhou2024lima,sharma2024critical}, they primarily address standard tasks like reducing harmfulness and increasing helpfulness, and do not assess fine-tuning methods for tasks requiring creativity, the focus of this work.

\textbf{RLHF Datasets.} Existing Reinforcement Learning with Human Feedback (RLHF) datasets, consisting of various responses to a prompt along with a preference ordering of those responses, have been critical for aligning existing AI systems to human preferences. We briefly review some of the most popular ones. 
Anthropic's HH-RLHF dataset~\citep{bai2022training} consists of chosen and rejected texts focusing on helpfulness and harmlessness. Stanford's SHP Dataset \citep{ethayarajh2022understanding} and Stack Exchange preferences dataset~\cite{askell2021general} have aggregated questions and answers along with their ratings from various online platforms. 
OpenAI's summarization dataset~\citep{hu2022sum} includes rankings of paired answers derived from human evaluations of text summaries. 
The data comes from a variety of sources, such as news articles and scientific papers, where human annotators compare the quality, coherence, and relevance of two different AI-generated summaries for the same text. 
The WebGPT comparisons~\citep{nakano2021webgpt} offer a dataset of human comparisons of AI-generated web search results, emphasizing the importance of high-quality, relevant information retrieval. 
Finally, we mention the Nectar dataset~\citep{starling2023}, which consists of a large series of prompts along with a list of five answers generated by various LLM's along with a ranking of these prompts by GPT-4.

\textbf{Humor in LLMs.} Several recent works have studied humor capabilities of large language models, in addition to the ones studying the New Yorker Caption Contest. Concurrent to our work, \citet{zhong2024let} also studies humor in a multi-modal setting, focusing on a different humor game Oogiri. Their experiments also suggests that existing chain of thought techniques are insufficient for LLMs to generate and understand humor. Similarly, \citet{jentzsch2023chatgpt} also shows GPT-3 still lacks humor abilities despite the good performance on other factual knowledge benchmarks. Furthermore, several works have focused on LLMs' capabilities in understanding humor, including humor detection~\citep{de2024thinc}, puns~\citep{xu2024good}, and humor explanation~\citep{chen2024talk}.

\section{New Yorker Caption Contest} \label{sec:contest}
\begin{figure*}
\begin{minipage}{0.5\linewidth}
    \includegraphics[width=\linewidth]{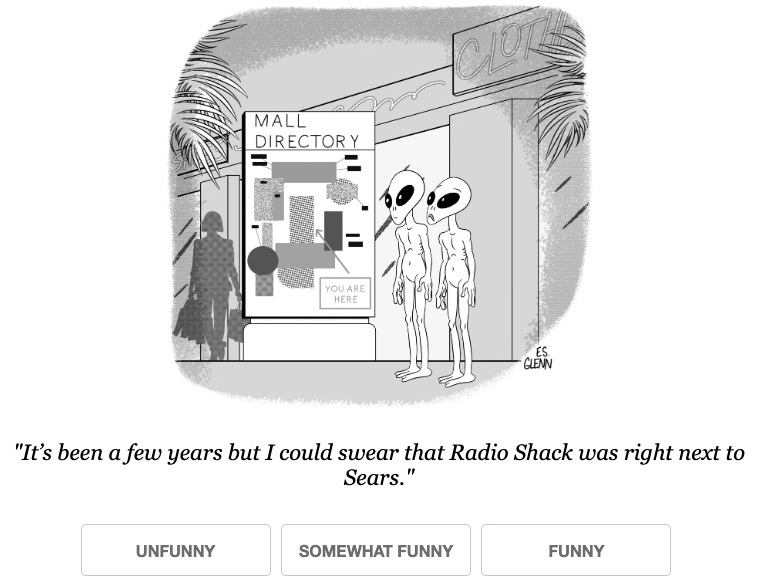}
    \captionof{figure}{Example voting page for contest 895}
    \label{fig:voting}
\end{minipage}
\quad
\begin{minipage}{0.45\linewidth}
    \captionof{table}{Dataset statistics}
    \begin{tabular}{ l c } 
      \toprule
      Number of contests & 365\\
      Number of cartoons & 365\\
      Average \#captions/contest & 6044\\
      STD \#captions/contest & 1794\\
      Total number of ratings & 284,183,913\\
      Average \#ratings/contest & 778,586 \\
      STD \#ratings/contest & 325,156\\  
      Max \#ratings/contest & 2,249,813\\
      Min \#ratings/contest & 31,173\\
      Average rating & 1.214$ (\pm$0.12)\\
      Top 10 average rating & 1.824 ($\pm$0.15)\\
      \bottomrule
    \end{tabular}
    \label{tab:stats}
\end{minipage}
\end{figure*}

Every week The New Yorker publishes an uncaptioned cartoon and solicits humorous captions from its readers through their website. 
The cartoon editors then review this list of captions and choose the top three funniest ones according to their judgement. The contest began in 2005, and at the time this work was written, there have been roughly 900 contests. For the last eight years, starting with contest 530, the New Yorker has utilized an online crowdsourced rating system (see Figure~\ref{fig:voting}) where users are presented with captions and can rate whether the caption is funny (a reward of 3), somewhat funny (a reward of 2), or unfunny (a reward of 1). 
Each week a large number of captions are submitted (on average more than 6,000).
These captions are first filtered by the New Yorker's editorial staff to remove captions that are not humorous or include personal information and/or offensive content, and then are sent to the crowdsourcing platform for large-scale rating.
%many of which are not particularly funny and can be filtered out efficiently. Moreover, captions including personal information and/or offensive content are filtered out. 
Finally, the New Yorker editors make their final decisions based on the crowdsourced ratings.

The rating process utilizes a multi-armed bandit-based algorithm, namely a UCB-variant (see~\cite{jamieson2015next,tanczos2017kl} and Appendix~\ref{apx:caption_contest} for details), to present users with higher-performing captions more frequently in order to efficiently identify the best caption. 
Additionally, since many of the captions are unfunny, this keeps the rating engaging by presenting users interesting captions to rate compared to random sampling. 
On average the contest receives close to 780,000 ratings per week. The top 5\% of captions receive an average of 821 ratings, and the bottom 50\% of captions receive around 85 ratings. 

The crowdsourced voting system for the New Yorker Caption Contest (NYCC) has resulted in an extensive dataset on human preferences and is a key contribution of this work. The dataset can be accessed at \url{https://huggingface.co/datasets/yguooo/newyorker_caption_ranking}. It consists of the cartoons, captions, and ratings for each one of 365 contests from contests 530 to 895. It provides an extensive labeled dataset on humor for researchers across multiple domains to study. In the related works, we describe some other works that have utilized this dataset. See Table~\ref{tab:stats} for more dataset statistics.

%and ensures higher statistical precision on truly better perfroming arms. 
%- keep the voting process interesting, - have statistical efficiency at the top

%Contests 530-890 (361 in total).

%In fact, it is 358 in total. Train data include the Jack Hessel Data after 530, 140 in total, together with the newly collected data from 761 to 890, 127 in total. Our test data include 91 contests, consisting of the test contests (47 in total) and the validations contests (44 in total) from the Jack Hessel dataset starting from contests 530. In total, we includes 358 contests. 

% \todo{Lalit and Jifan: Description of the new yorker caption contest.}

% \todo{Lalit and Jifan: What we provide in the dataset :
% \begin{enumerate}
%     \item Human submissions.
%     \item Ratings on each caption: funny, somewhat funny and not funny.
%     \item Scoring method: funny 3 points, somewhat 2 points, unfunny 1 point.
%     \item An example screenshot of leaderboard.
%     \item GPT4-V generated image description.
% \end{enumerate}}
  
% \subsection{Ranking and Crowdsourcing Method}
% \todo{Lalit:
% Screenshot of the voting page. Description of voting process. Description of statistical significance and algorithm used to obtain rankings.

% Which part of the rankings are more statistically reliable? Limitations?}

\section{HumorousAI Benchmark: Funny Cartoon Caption Generation}
In this section, we establish a benchmark method for evaluating the ability of large language models to generate funny captions. We start by describing the tasks in Section~\ref{ssec:task} followed by our proposed evaluation methods described in Section~\ref{ssec:eval}. Lastly, in Section~\ref{ssec:finetune}, we give a brief overview of the various finetuning methods we explore in this paper.

\subsection{Task} \label{ssec:task}
\begin{table}[ht]\small
    \centering
    \caption{\textbf{Evaluation reliability measure}: Ranking accuracy of captions ranked \#1-10 vs captions ranked \#1000-1009 averaged over 200 pairs. See Appendix~\ref{apx:description_generation} for details on how the cartoon descriptions are generated.}
    \label{tab:ranking}
    \begin{tabular}{c l c c}
        \toprule
        \textbf{Comparison Method} & \shortstack{Evaluator} & Description/Image & Ranking Accuracy(\%)  \\
        \midrule
        \multirow{8}{*}{\textbf{Pairwise}}
        & Human (worker) & GPT4o-vision & 61.67$\pm$3.45\\
        & Human (worker) & Cartoon Image & 60.79$\pm$3.46\\
        & GPT4-Turbo-vision & Cartoon Image & 61$\pm$3.46 \\
        & GPT4o-vision & Cartoon Image & 60.5$\pm$3.47\\
        & GPT4o & GPT4o-vision & 65$\pm$3.38\\
        & \textbf{GPT4-Turbo} & \textbf{GPT4o-vision} & \textbf{67$\pm$3.33}\\
        & GPT4-Turbo & GPT4-vision & 66$\pm$3.36 \\
        & GPT4-Turbo & \citet{hessel2022androids} & 66.5$\pm$3.35\\
        \midrule
        \multirow{7}{*}{\textbf{\shortstack{Group\\(Overall)}}} 
        & Human (worker) & GPT4o-vision & 59.23$\pm$1.45\\
        & Human (worker) & Cartoon Image & 57.5.42$\pm$1.37 \\
        & Human (expert) & Cartoon Image & 94.28$\pm$2.79\\
        & GPT4-Turbo-vision & Cartoon Image  & 63$\pm$3.42\\
        & GPT4o-vision & Cartoon Image  & 74$\pm$3.11\\
        & GPT4-Turbo & GPT4o-vision  & 73$\pm$3.15\\
        & GPT4-Turbo & GPT4-vision & 74$\pm$3.11\\
        & \textbf{GPT4-Turbo} & \textbf{\citet{hessel2022androids}} & \textbf{77.5$\pm$2.96}\\
        \midrule
        \multirow{5}{*}{\textbf{\shortstack{Group\\(Best Pick)}}} 
        & Human (worker) & GPT4o-vision & 56 $\pm$ 2.22 \\
        & Human (worker) & Cartoon Image & 63.66$\pm$1.96 \\
        & \textbf{GPT4o-vision} & \textbf{Cartoon Image}  & \textbf{70.5$\pm$3.23} \\
        & GPT4-Turbo & GPT4o-vision  & 61.5$\pm$3.45 \\
        & GPT4-Turbo & \citet{hessel2022androids} & 60$\pm$3.47 \\
        \bottomrule
    \end{tabular}
\end{table}

We focus on the cartoon captioning task in this paper, where a model is given the information about the cartoon and is asked to generate funny captions about it. Specifically, we evaluate both multimodel large language models (MLLMs) and language-only models (LLMs). For MLLMs, we provide the raw cartoon images. For language-only models, we instead provide the descriptions and object entities of the cartoons. The text format of these descriptions are either written by human~\citep{hessel2022androids} or generated by MLLMs by given the images (see Appendix~\ref{apx:description_generation} for details). See~\cref{tab:description} for the example descriptions.

We hold out a set of $91$ out of the 358 contests for evaluation by an evaluator (see Section~\ref{ssec:eval}). For each contest and its corresponding cartoon, we ask the language model to generate ten captions. This group of ten captions is then compared against four groups of past human submissions by the evaluator. For each contest, the four groups are captions ranked \#1-10, \#200-209, \#1000-1009 and the ten captions that received median ranking. The evaluations are conducted along three dimensions:
\begin{enumerate}
    \item \textbf{Overall comparison}: In this setting, the evaluator compares the overall funniness of the group of model-generated captions against each group of contestant-submitted captions. Win rates of the model-generated captions will be reported in Section~\ref{sec:experiments} and Table~\ref{tab:eval_accs}.

    \item \textbf{Best pick comparison}: We ask the evaluator to first pick the funniest caption from each of the two groups and then choose the funnier caption accordingly. Win rates are reported similarly to above.

    \item \textbf{Caption diversity}: We measure the diversity of captions within each group of captions either generated by language models or submitted by human contestants in the past. Similarly to the study~\citep{kirk2023understanding} on measuring the output diversity for non-creative tasks (summarization and instruction following), we use the expectation-adjusted distinct N-grams (denoted as \textbf{Average EAD})~\citep{li2015diversity} and the Sentence-BERT embedding cosine similarity (denoted as \textbf{SBERT})~\citep{reimers2019sentence} to measure the per-contest diversity. \textbf{Average EAD} measures the token-level similarity of the generated captions, while \textbf{SBERT} measures the semantic-level similarity. We do not use the NLI diversity from~\citep{stasaski2022semantic} as it is conversation-specific. 
\end{enumerate}
Our evaluation primarily focuses on comparing groups of captions since evaluation reliability can be significantly improved as we now discuss below. 

\subsection{Evaluation Method} \label{ssec:eval}
\begin{quote}
    \textit{Humor is notoriously subjective. Humans cannot infallibly predict what other humans will find funny. If they could, no joke would ever fall flat. We just do the best we can, always hoping we can do better. Likewise for these models.} \hfill--Bob Mankoff, former cartoon editor of The New Yorker
\end{quote}

In this section, we aim to find a comparably reliable evaluation method for judging model-generated captions against human submissions. We experimented with various versions of GPT-4 and also human evaluations from Prolific \citep{palan2018prolific}. This task has been studied widely before within the context of humor~\citep{shahaf2015inside,radev2015humor,king2013random,hessel2022androids}. However, unlike these previous studies that only evaluate two candidate captions at a time (denoted by \textbf{Pairwise}), we introduce the novel group comparison techniques for evaluation (denoted by \textbf{Group Overall} and \textbf{Group Best Pick}). As described in Section~\ref{ssec:task}, we compare groups of ten captions from different sources, such as human submissions from different ranking levels, or captions generated by different language models. 
To measure the reliability of different evaluators, as reported in Table~\ref{tab:ranking}, we compare their accuracy in judging human-submitted captions from top \#10 versus \#1000-1009 across $200$ different contests. 
For the \textbf{Pairwise} comparisons, we uniformly at random choose one caption from each of the two groups, which exactly corresponds to the \emph{ranking} task proposed by \citet{hessel2022androids}. 
For group comparisons, we provide all ten captions from each group to a single query to an LLM/human rater. 
The detailed prompts can be found in Appendix~\ref{apx:prompts} for various language models. 
All of the prompts for evaluation utilize the 5-shot in-context prompting technique, which provides five caption comparison examples from other contests before asking the model to rank the pair/groups of captions for the given cartoon.

As shown in Table~\ref{tab:ranking}, language models are generally more accurate in detecting the higher-ranked favorable captions in a group comparison paradigm compared to the pairwise paradigm. 
These models also outperform average humans (crowd workers) in judging the funniness across all three comparison settings. Notably, in the overall group comparisons we also included evaluations from a human expert (the former cartoon editor for The New Yorker). The expert significantly outperforms all other evaluators (AI and human), exposing a significant gap between human experts and SOTA AI systems in this domain. Also, the group comparisons are somewhat more challenging for crowd workers than pairwise comparisons, but group comparisons make the language model evaluations much more reliable and accurate. Further details about the evaluations can be found in Appendix~\ref{apx:additional_setup}.

In conclusion, we establish two benchmark evaluation methods for the rest of this paper: \textbf{Group Comparison (Overall)} using GPT4-Turbo as evaluator with descriptions from \citet{hessel2022androids} and \textbf{Group Comparison (Best Pick)} using GPT4o-vision as evaluator with raw cartoon images.

\newcommand{\rst}{r^\star}
\newcommand{\Dpref}{\mathcal{D}_{\mathrm{pref}}}
\newcommand{\rhat}{\widehat{r}}

\subsection{Alignment Finetuning Methods}\label{ssec:finetune}
In our study, we compare the performance of a 0-shot model (with standard and Best-of-N sampling) to that of an SFT finetuned model, an RLHF finetuned model, and a DPO finetuned model. We briefly outline these methods here, and refer the reader to \citep{christiano2017deep,bai2022training,ouyang2022training,rafailov2024direct} for further details. In all cases, we adopt the implementation from the TRL package~\citep{von_Werra_TRL_Transformer_Reinforcement}.

\textbf{Supervised Finetuning (SFT):} 
SFT assumes access to a dataset $\mathcal{D}_{\mathrm{sft}} = \{ (x^{(i)},y^{(i)}) \}_{i=1}^{N}$ of prompt-completion pairs, where $y^{(i)}$ is assumed to be an ``expert'' completion for prompt $x^{(i)}$. SFT then tunes the weight of the base model to maximize the likelihood of completions $y^{(i)}$ given prompt $x^{(i)}$.

\textbf{Reinforcement Learning from Human Feedback (RLHF):}
RLHF assumes access to a preference dataset $\Dpref = \{ (x^{(i)}, y_w^{(i)}, y_l^{(i)}) \}_{i=1}^M$, where $x^{(i)}$ is a prompt, and $y_w^{(i)}, y_l^{(i)}$ two possible completions to $x^{(i)}$, where $y_w^{(i)}$ is preferred over $y_l^{(i)}$. 
RLHF assumes these preferences are consistent with an (unknown) reward function $\rst$, typically assumed to follow the Bradley-Terry model \citep{bradley1952rank}. It first trains a reward model $\rhat$ on $\Dpref$, and then finetunes the base language model to maximize $\rhat$, typically running PPO \citep{schulman2017proximal} and regularizing the training to ensure it does not deviate significantly from the SFT model.\loose

\textbf{Direct Preference Optimization (DPO):}
DPO operates under the same assumptions as RLHF, but skips the reward modeling step entirely, and instead finetunes the base language model on $\Dpref$ directly, tuning it to produce next-token likelihoods with orderings consistent with $\Dpref$.

\textbf{Best-of-N Sampling (BoN):}
Best-of-N sampling does not modify the weights of the base model. Instead, it samples $N$ completions from the base model for any prompt $x$, and chooses the completion with the highest reward, as quantified by the reward $\rhat$ obtained from the RLHF reward-learning step.

%Both DPO and RLHF assumes these preferences are consistent with an (unknown) reward function $\rst$, typically assumed to follow the Bradley-Terry model \citep{bradley1952rank}. RLHF first trains a reward model $\rhat$ on $\Dpref$, and then finetunes the base language model to maximize $\rhat$, typically running PPO \citep{schulman2017proximal} and regularizing the training to ensure it does not deviate significantly from the SFT model. DPO skips the reward modeling step entirely, and instead finetunes the base language model on $\Dpref$ directly, tuning it to produce next-token likelihoods with orderings consistent with $\Dpref$.

\textbf{Preference Dataset Construction:}
In our setting, we take $\mathcal{D}_{\mathrm{sft}}$ to be a dataset of cartoon-caption pairs, where the captions $y^{(i)}$ are drawn at random from the entire training set of captions for cartoon $x^{(i)}$. $\Dpref$ is constructed by taking a cartoon $x^{(i)}$ and then two captions $y_w^{(i)}$ and $y_l^{(i)}$, where $y_w^{(i)}$ is set to a caption with a higher human rating then $y_l^{(i)}$. Specifically, we sample the pair to be at least 3 standard deviation apart from each other, i.e. 
\begin{equation}\label{eq:3_std}
\text{Rating}(y_w^{(i)}) - \text{Rating}(y_l^{(i)}) \geq 3 \cdot \sqrt{\text{STD}(y_w^{(i)} )^2+\text{STD}(y_l^{(i)})^2},
\end{equation}
where $\text{Rating}(y)$ is the average score of caption $y$ from human raters according to rewards defined in Section~\ref{sec:contest} (note this is different from the rewards from the reward model of RLHF). $\text{STD}(y)$ is the corresponding standard deviation of scores from human raters.

\section{Experiments} \label{sec:experiments}
In this study, we evaluate the performance of caption generation. We experiment with two open-source large language models, Mistral 7b Instuct (mistralai/Mistral-7B-Instruct-v0.1)\citep{jiang2023mistral} and the multimodal model LLaVa 7b (llava-hf/llava-v1.6-mistral-7b-hf)\citep{liu2024visual} finetuned with methods in Section~\ref{ssec:finetune}. We also evaluate state-of-the-art close-sourced models including GPT4o and Claude 3 Opus. See Appendix~\ref{apx:exp_details} for more details. Our code is available at \url{https://github.com/yguooo/cartoon-caption-generation}.

\begin{table}[t]\small
\caption{Evaluation of captions generated by various language models. We utilize group comparison strategies mentioned in Section~\ref{ssec:eval}. The generated captions are compared against four groups of human contestant entries at different ranking levels. Win rates are based on $91$ held-out cartoons.}\label{tab:eval_accs}
\scalebox{.98}{
\begin{tabular}{lcccccccc}
\toprule
 & \multicolumn{4}{c}{Overall Win Rate (\%) $\uparrow$} & \multicolumn{4}{c}{Best Pick Win Rate (\%) $\uparrow$} \\
\cmidrule(r){2-5} \cmidrule(l){6-9}
\shortstack{Generated Caption\\ Model}  & Top 10 & \shortstack{\#200-\\\#209} & \shortstack{\#1000-\\\#1009} & \shortstack{Contestant\\Median} & Top 10 & \shortstack{\#200-\\\#209} & \shortstack{\#1000-\\\#1009} & \shortstack{Contestant\\Median} \\ \midrule
LLaVA & 3.85 & 2.20 & 4.40 & 13.19 &   2.75     &  6.59   &  4.95  &  12.64 \\ 
LLaVA SFT & 2.75 & 3.30 & 7.14 & 17.03 &  2.20      & 4.95    & 6.59   & 10.99  \\ 
Mistral-7B 0-Shot & 4.95 & 8.79 & 11.54 & 25.82 &  1.65    &  1.65   & 3.85   & 12.64  \\
Mistral-7B BoN   & 6.59 & \textbf{16.48} & \textbf{21.43} & \textbf{35.71} &  1.65    & 2.20    & 3.30 & 10.44   \\
Mistral-7B SFT & 3.85 & 4.40 & 7.14 & 14.29  &   0.55    &   2.20  & 1.65  &  8.24    \\ 
Mistral-7B RLHF & 8.79 & 9.34 & 11.54 & 24.73 &  2.20 & 3.30 & 8.24 & 13.19\\
Mistral-7B DPO & \textbf{9.34} & 13.74 & 17.58 & 31.32 &    \textbf{10.44}   & \textbf{15.93}   &  \textbf{14.29}  & \textbf{30.22}     \\
% Mistral-7B DPO BoN &\textbf{11.54} & 9.89 & 16.48 & 30.77 & 7.14 & 11.54 & \textbf{14.29} & 28.57     \\
\midrule
GPT-3.5 Turbo & 33.52 & 52.75 & 62.09 & 76.92 & 23.63 & 46.7 & 48.35 & 70.88 \\
GPT-4o & 44.51 & 69.23 & 79.12 & 86.81 & 42.86 & 59.89 & 73.63 & 79.67 \\
GPT-4o Vision & 42.31 & 63.74 & 76.92 & 85.16 & \textbf{47.80} & \textbf{65.93} & \textbf{79.67} & \textbf{85.71} \\
Claude-3-Opus & \textbf{54.40} & \textbf{70.88} & \textbf{81.87} & \textbf{88.46} & 40.11 & 59.89 &63.74 &79.67 \\
\bottomrule
\end{tabular}
}
\end{table}

\subsection{Experimental Results}
In \cref{tab:eval_accs}, we report the result for pretrained and finetuned model generations evaluated by GPT models. In \cref{tab:human_eval}, we ask human workers and expert to evaluate the captions generated by SOTA models. Below, we document some of our findings and research questions they inspire.

\textbf{MLLMs vs LLMs.} Surprisingly, language-only models such as the pretrained Mistral model outperform the multimodal LLaVa model that has access to the entire cartoon images. Similarly, for overall group comparison, GPT-4o is also preferred over GPT-4o with vision. To further investigate this issue, we conducted more experiments and obtained the following results:
\begin{enumerate}
    \item GPT4-Turbo as evaluator given GPT4o descriptions (as reported in Table 2). Accuracy: 67\%.

    \item GPT4-Turbo-vision as evaluator given cartoon and GPT4o descriptions. Accuracy: 60.5\%.

    \item GPT4-Turbo-vision as evaluator given a blank image and GPT4o descriptions. Accuracy: 61.5\%.
\end{enumerate}
For bullet points 2 and 3 above, we are running the exact same model, the only difference is that one has access to the cartoon + text description, while the other has access only to the text description, thus isolating the effect of the image on the generation quality. We find that the visual element integration into the LLMs is negatively biasing the model’s accuracy. This observation is also consistent for overall and best pick group comparisons. Since giving a blank image also hurts performance compared to bullet point 1, GPT4 without vision, it is unlikely that the performance of the vision model is dragged down by the visual understanding capabilities.

One possible reason for the above observation is that the training corpus for multimodal LLMs can be much less diverse than the training corpus for the LLM. For example, LLaVa is only trained on a small multi-modal instruction following dataset ($\sim$80K unique images)~\citep{liu2024visual}, whereas generic LLMs like Mistral or Llama are trained on much larger dataset. Overall, these findings suggest there is still much research to be done in better integrating multi-modal capabilities into large language models.
\begin{center}
\fbox{
\begin{minipage}{.95\linewidth}
    Proposed Research Question \#1: The multimodal large language models still underperform their language-only counterparts in caption generation. Can  the vision-language integration in MLLMs be further improved to close this gap?
\end{minipage}
}
\end{center}

\textbf{Finetuning Open Source Models.} 
We observe that supervised fine-tuning hurts the model performance in the humor generation task in general. We believe this is primarily because we are aligning to captions in the top $1000$, most of which are not particularly funny. However, we note this is an important step before RLHF and DPO training, as it trains the models to generate captions in the correct format.
We also find that BoN sampling is able to substantially increase the Overall Win Rate metric, but falls short on the Best Pick Win Rate, which suggests the reward model is favoring a small set of good captions, but none of which generates particularly outstanding captions. We also observe in the next section that BoN indeed results in a less diverse group of generations.

As compared to BoN, running RLHF on the same reward model is unable to achieve as high a level of performance. As we show in~\cref{apx:additional_results}, running PPO does indeed yield generations with higher reward score as given by the reward model, and as our BoN results indicate, filtering captions based on their reward does give better performance. 
This suggests that, while our reward model is able to effectively filter generations, tuning a model to maximize it does not necessarily lead to improved performance.
We hypothesize that this is due to the complex nature of humor and the potential for out-of-distribution generations when running RLHF. While our reward model may effectively rank captions within a set of reasonable and in-distribution captions (for example those generated by the 0-shot model), small deviations from the training distribution could lead to an erroneous reward signal. Furthermore, for tasks such as humor generation very subtle changes (for example, minor changes in word choice) can drastically change how humorous a caption is---the distribution of humorous captions is extremely sensitive. Together, we believe these phenomenon make it challenging for PPO to effectively finetune the weights to obtain significantly more humorous generations.
%We hypothesize that this is due to the complex nature of humor and the potential for out-of-distribution generations when running RLHF. While our reward model may effectively rank captions within a set of reasonable and in-distribution captions (for example those generated by the 0-shot model), small deviations from the training distribution in RLHF could lead to an erroneous reward signal. We believe this is particularly an issue in humor generation, where very subtle changes (for example, minor changes in word choice) can drastically change how humorous a caption is---the distribution of humorous captions is extremely sensitive. 
\begin{center}
\fbox{
\begin{minipage}{.95\linewidth}
    Proposed Research Question \#2: Can we train a reward model able to better capture humor? Can RLHF still be effectively applied to settings where the distribution of correct responses is highly sensitive?\loose
\end{minipage}
}
\end{center}

In contrast to RLHF, DPO does yield a significant increase over the 0-shot model for the Best Pick Win Rate metric. 
Note that DPO only optimizes the model on offline preference data and, as such, does not require an evaluation of any out-of-distribution samples. We hypothesize that, in settings such as humor generation where the desired distribution is extremely sensitive, this could lead to better performance, as it avoids the aforementioned issue where RLHF may quickly drift to producing out-of-distribution samples, for which the reward signal is erroneous. 

\begin{center}
\fbox{
\begin{minipage}{.95\linewidth}
    Proposed Research Question \#3: Does DPO lead to better in-distribution generation, and produce a model more effectively able to match the distribution of the finetuning data?
\end{minipage}
}
\end{center}

\begin{wraptable}{r}{.5\linewidth}
    \caption{Rate of Claude-3-Opus generated captions preferred over Human Top 10.}
    \centering
    \begin{tabular}{lc}
    \toprule
    Evaluator & Preference Rate \\
    \midrule
     Human (expert)    & \textbf{1.6\%}\\
     Human (worker)    & \textbf{35.4\%} \\
    \bottomrule
    \end{tabular}
    \label{tab:human_eval}
\end{wraptable}
\textbf{Human Evaluation.}
We also ran a human evaluation using six workers from Prolific~\citep{palan2018prolific} along with a humor expert (a former New Yorker editor) to understand how often people preferred caption generations from Claude vs top 10 ranked captions generated by humans. We find that people only prefer Claude's generations 34\% of the time. Our expert preferred Claude's generation only 1.6\% of the time. He said, \textit{``I think I preferred human captions because from my ``expert" vantage point they were better phrased and more concise even independent from being funny. At this point AI tends to be too verbose in almost any task but, for me that is a liability when it comes to creating a good caption."}

This suggests that, though SOTA LLMs can generate a diverse set of funny captions, there remains a significant gap in their humor and creativity when judged from the perspective of human experts.\loose

\textbf{Example Generation and Qualitative Analysis.} As shown in~\cref{tbl:example_generation}, we provide some generation samples for the cartoon in Figure~\ref{fig:voting}. Indeed, we see that LLMs generally produce longer and more verbose captions than top human ones. Moreover, we generate multiple additional captions with GPT-4o-vision and Claude-3-Opus for the cartoon in Figure~\ref{fig:voting}. Below, we make a qualitative analysis around the shortcoming of these generated captions from SOTA LLMs. 

\begin{itemize}
    \item \textbf{Missing visual details resulting in LLMs generating out-of-context captions.} As an example, GPT-4o-vision generated another caption of \emph{``So the alien abduction statistics were right. Malls are the prime hunting grounds!''} This caption does not match the cartoon though, since the aliens look missing and worried. Their bodies look skinny and weak. In other words, they don’t seem to be here to hunt humans.

    \item \textbf{Many generated captions are forms of word/phrase modification and creation.} An example of this caption is \emph{``Do they have a Black Hole Friday sale?''}, also generated by GPT4o-vision. Another example from Claude-3-opus is \emph{``I don't see 'Invasion Supplies' listed anywhere...''} Both of these captions are inventing new words and phrases to make the caption funny. While these can be somewhat funny, they are usually not rated highly by the New Yorker audiences.

    \item \textbf{Winning captions tend to appeal to readers with multiple interpretations through different lenses.} LLMs currently lack the ability to produce such captions. The New Yorker’s editorial pick of the final caption was \emph{``Oh sure, now you look at a map.''} While this caption makes fun of the aliens blaming each other, it also references the experience of a driver missing the direction but claiming they knew the way. On the contrary, GPT4o-vision generated the caption \emph{``We travel light-years, and we still need directions!''}, which despite making fun of the characters in the cartoon with the same concept, lacks the relatability from an additional perspective.
\end{itemize}
Overall, we think the GPT4 models are capable of generating humorous content. However, to rank among the top requires much deeper understanding of cultural references and more steps of reasoning before arriving at a high quality caption.

\subsection{Diversity Evaluation}\label{sec:diversity}

We evaluated the token-level and semantic-level diversity of the generation with results given in~\cref{tab:eval_diversity}. We found the Average EAD and SBERT share the same trend when the base model is the same. Within the human generated caption group, we noticed that their diversity scores are very similar under both metrics. And the human generated texts regardless of their funniness are much more diverse than any model-generated captions. 

For pretrained models, the commercial models like GPT, Claude-3 generally outperform the open-source model, like Mistral or LLaVa, in terms of diversity. Introducing the SFT and PPO procedure can moderately improve the diversity metrics for the Mistral model. This is in contrast to the findings of \citep{kirk2023understanding}, which observed the opposite effect, that RLHF reduced diversity in regular text generation tasks. We also found that running DPO can yield a significant increase in the diversity of the model generations as compared to any other method. We hypothesize that this may be due to our finetuning dataset: for each cartoon, we run DPO with a variety of human-generated captions and it therefore learns not to prefer a single caption or type of caption, but a diversity of captions.
\begin{center}
\fbox{
\begin{minipage}{.95\linewidth}
    Proposed Research Question \#4: DPO exhibits surprisingly good diversity metrics as compared to PPO and SFT. Does the data diversity used for finetuning explain this, or are other mechanisms at play?\loose
\end{minipage}
}
\end{center}

%The most interesting result is that DPO can significantly improve the model output diversity, and achieve even better diversity than the commercial language models. These results provide a different observations in the creative generation tasks compared to~\citep{kirk2023understanding}, which observed a reduced per-input diversity for RLHF. Our hypothesis is that the implicit reward training in DPO allows the reward training be adaptive with respect to inputs than fixing the reward at the start.  \todo{Andrew probably has a better idea for explaining it.}

% \textbf{Research Question 4: DPO exhibits surprisingly good diversity metrics beyond PPO and SFT. Studies into their internal mechanism can potentially lead to improved model design for creative generation tasks.}

% \todo{Yang \& Jifan: Discussion about splits, experiment setup, evaluation metrics (diversity, accuracy@diversity).}

% \yang{I think changing p can possibly align the diversity score for different model. But I can draw the tradeoff curve when I vary p. }

% \textbf{Add diversity vs accuracy tradeoff curve.}

% \todo{Finding: Different KL penalty incurs different behaviors.}
\begin{figure*}
\renewcommand{\arraystretch}{1.2} % Default value: 1
\begin{minipage}{.49\linewidth}\small
\captionof{table}{Example caption generations for \\contest \#895 (cartoon in \cref{fig:voting})}\label{tbl:example_generation}
\centering
\begin{tabularx}{\textwidth}{l|X}
\toprule
Mistral-7B 0-shot &  \texttt{When your GPS leads you to the wrong galaxy.} \\
Mistral-7B BoN  &   \texttt{What do you call it when aliens invade your favorite mall? A takeover by outer space retailers!}                                 \\
Mistral-7B DPO       &    \texttt{I just assumed we were the only ones who knew how to pronounce ``H\&M''.}                                \\
\midrule
GPT4-o-vision       & \texttt{We travel light-years, and we still need directions!}                                    \\
Claude-3-Opus       &  \texttt{Let's hit the food court first. I'm craving some Jupiter fries.}                                 \\
\midrule
Human Winner & \texttt{Do you think death rays would be considered electronics or sporting goods?}\\
\bottomrule
\end{tabularx}
\end{minipage}
\renewcommand{\arraystretch}{1.3} % Default value: 1
\begin{minipage}{.49\linewidth}\small
\captionof{table}{Diversity evaluation on the generated captions. We use the expectation-adjusted distinct N-grams (Average EAD)~\citep{li2015diversity} and the Sentence-BERT embedding cosine similarity (SBERT)~\citep{reimers2019sentence} to measure the per-contest diversity on the token level and semantic level. }\label{tab:eval_diversity}
\centering
\begin{tabular}{lcc}
\toprule
Caption Source & \shortstack{Average\\EAD $\uparrow$}     & SBERT $\uparrow$          \\
\midrule
Human (Top 10)    & 0.9456          & 0.7452          \\
Human (\#200-\#209)     & 0.9564          & 0.7496          \\
Human (\#1000-\#1009)      & 0.9608 & 0.7522 \\
Human (Median)  & 0.9597          & 0.7489          \\
\midrule                                     
LLaVA & 0.8986          & 0.5220          \\
LLaVA SFT & 0.9002          & 0.5173          \\ 
\midrule
Mistral-7B Instruct 0-Shot & 0.9037          & 0.5349          \\
Mistral-7B Instruct BoN     & 0.8663          & 0.4868          \\
Mistral-7B Instruct SFT & 0.9043          & 0.5806          \\
% Mistral-7B DPO + BoN & 0.9198           & 0.7127           \\
Mistral-7B Instruct RLHF & 0.9006          & 0.5994          \\
Mistral-7B Instruct DPO & 0.9206          & 0.7075          \\
\midrule
GPT4-o  & 0.9602          & 0.5789          \\
Claude-3-Opus & 0.9533          & 0.6813 \\
\bottomrule
\end{tabular}
\end{minipage}
\end{figure*}

\section{Future Work and Societal Impact}\label{sec:future_work}
This paper opens a suite of research problems and challenges going forward and we are excited to continue working on multiple directions of future work.

\textbf{Improving creativity in LLM generation.} While LLMs are largely applauded for their creativity today, our experiments reveal there is still a significant gap between top human generated content and SOTA LLMs and MLLMs, especially when judged by an expert. We believe addressing the proposed research questions can not only improve funny caption generation, but also improve existing models on the creative generation tasks in general.

\textbf{Gamified evaluation of AI generated captions by a crowd.} As the nature of the funny cartoon captioning task is an engaging game by nature, we plan on building an AI versus Human battle ground rating game. Our envisioned game will allow users to submit their own captions. During rating, participants are presented with two sets of captions from different sources (human vs human, human vs AI and AI vs AI). This also provides us a more reliable system for evaluating new captions on new cartoons. At the same time, researchers are encouraged to submit AI model entries to test out their latest model/alignment methods.

\textbf{Humor vs offensiveness tradeoff.} Optimizing for humor abilities may result in increasing offensiveness and toxicity of model generated content. We believe an important next step is to study the  challenge of balancing humor with potential offensiveness. As the boundary between humorous and offensive are often blurred, the subjective nature of humor and cultural sensitivities needs to be futher studied to ensure AI models align with human values.

\begin{ack}
{\bf Funding (financial activities supporting the submitted work):} This work was partially supported by the NSF projects 2023239 and 2112471.  
% \rob{first project is IFDS, which supports a bunch of us at Madison and Washington.  second is AI Edge.} 

{\bf Competing Interests (financial activities outside the submitted work):} R. Mankoff is formerly the cartoon editor of The New Yorker magazine and currently works with CartoonStock, a commercial cartoon company.  L. Jain and R. Nowak formerly provided crowdsourcing services to The New Yorker magazine.
\end{ack}

\newpage
\bibliography{reference}

\begin{thebibliography}{68}
\providecommand{\natexlab}[1]{#1}
\providecommand{\url}[1]{\texttt{#1}}
\expandafter\ifx\csname urlstyle\endcsname\relax
  \providecommand{\doi}[1]{doi: #1}\else
  \providecommand{\doi}{doi: \begingroup \urlstyle{rm}\Url}\fi

\bibitem[Askell et~al.(2021)Askell, Bai, Chen, Drain, Ganguli, Henighan, Jones, Joseph, Mann, DasSarma, et~al.]{askell2021general}
Amanda Askell, Yuntao Bai, Anna Chen, Dawn Drain, Deep Ganguli, Tom Henighan, Andy Jones, Nicholas Joseph, Ben Mann, Nova DasSarma, et~al.
\newblock A general language assistant as a laboratory for alignment.
\newblock \emph{arXiv preprint arXiv:2112.00861}, 2021.

\bibitem[Auer(2002)]{auer2002using}
Peter Auer.
\newblock Using confidence bounds for exploitation-exploration trade-offs.
\newblock \emph{Journal of Machine Learning Research}, 3\penalty0 (Nov):\penalty0 397--422, 2002.

\bibitem[Azar et~al.(2024)Azar, Guo, Piot, Munos, Rowland, Valko, and Calandriello]{azar2024general}
Mohammad~Gheshlaghi Azar, Zhaohan~Daniel Guo, Bilal Piot, Remi Munos, Mark Rowland, Michal Valko, and Daniele Calandriello.
\newblock A general theoretical paradigm to understand learning from human preferences.
\newblock In \emph{International Conference on Artificial Intelligence and Statistics}, pages 4447--4455. PMLR, 2024.

\bibitem[Bai et~al.(2022{\natexlab{a}})Bai, Jones, Ndousse, Askell, Chen, DasSarma, Drain, Fort, Ganguli, Henighan, et~al.]{bai2022training}
Yuntao Bai, Andy Jones, Kamal Ndousse, Amanda Askell, Anna Chen, Nova DasSarma, Dawn Drain, Stanislav Fort, Deep Ganguli, Tom Henighan, et~al.
\newblock Training a helpful and harmless assistant with reinforcement learning from human feedback.
\newblock \emph{arXiv preprint arXiv:2204.05862}, 2022{\natexlab{a}}.

\bibitem[Bai et~al.(2022{\natexlab{b}})Bai, Kadavath, Kundu, Askell, Kernion, Jones, Chen, Goldie, Mirhoseini, McKinnon, et~al.]{bai2022constitutional}
Yuntao Bai, Saurav Kadavath, Sandipan Kundu, Amanda Askell, Jackson Kernion, Andy Jones, Anna Chen, Anna Goldie, Azalia Mirhoseini, Cameron McKinnon, et~al.
\newblock Constitutional ai: Harmlessness from ai feedback.
\newblock \emph{arXiv preprint arXiv:2212.08073}, 2022{\natexlab{b}}.

\bibitem[Bakker et~al.(2022)Bakker, Chadwick, Sheahan, Tessler, Campbell-Gillingham, Balaguer, McAleese, Glaese, Aslanides, Botvinick, et~al.]{bakker2022fine}
Michiel Bakker, Martin Chadwick, Hannah Sheahan, Michael Tessler, Lucy Campbell-Gillingham, Jan Balaguer, Nat McAleese, Amelia Glaese, John Aslanides, Matt Botvinick, et~al.
\newblock Fine-tuning language models to find agreement among humans with diverse preferences.
\newblock \emph{Advances in Neural Information Processing Systems}, 35:\penalty0 38176--38189, 2022.

\bibitem[Bradley and Terry(1952)]{bradley1952rank}
Ralph~Allan Bradley and Milton~E Terry.
\newblock Rank analysis of incomplete block designs: I. the method of paired comparisons.
\newblock \emph{Biometrika}, 39\penalty0 (3/4):\penalty0 324--345, 1952.

\bibitem[Burns et~al.(2023)Burns, Izmailov, Kirchner, Baker, Gao, Aschenbrenner, Chen, Ecoffet, Joglekar, Leike, et~al.]{burns2023weak}
Collin Burns, Pavel Izmailov, Jan~Hendrik Kirchner, Bowen Baker, Leo Gao, Leopold Aschenbrenner, Yining Chen, Adrien Ecoffet, Manas Joglekar, Jan Leike, et~al.
\newblock Weak-to-strong generalization: Eliciting strong capabilities with weak supervision.
\newblock \emph{arXiv preprint arXiv:2312.09390}, 2023.

\bibitem[Chakraborty et~al.(2024)Chakraborty, Qiu, Yuan, Koppel, Huang, Manocha, Bedi, and Wang]{chakraborty2024maxmin}
Souradip Chakraborty, Jiahao Qiu, Hui Yuan, Alec Koppel, Furong Huang, Dinesh Manocha, Amrit~Singh Bedi, and Mengdi Wang.
\newblock Maxmin-rlhf: Towards equitable alignment of large language models with diverse human preferences.
\newblock \emph{arXiv preprint arXiv:2402.08925}, 2024.

\bibitem[Chan et~al.(2024)Chan, Sun, Holt, and van~der Schaar]{chan2024dense}
Alex~J Chan, Hao Sun, Samuel Holt, and Mihaela van~der Schaar.
\newblock Dense reward for free in reinforcement learning from human feedback.
\newblock \emph{arXiv preprint arXiv:2402.00782}, 2024.

\bibitem[Chang et~al.(2024)Chang, Shan, Oertell, Brantley, Misra, Lee, and Sun]{chang2024dataset}
Jonathan~D Chang, Wenhao Shan, Owen Oertell, Kiant{\'e} Brantley, Dipendra Misra, Jason~D Lee, and Wen Sun.
\newblock Dataset reset policy optimization for rlhf.
\newblock \emph{arXiv preprint arXiv:2404.08495}, 2024.

\bibitem[Chen et~al.(2024{\natexlab{a}})Chen, Yuan, Liu, Liu, Guan, Guo, Peng, Liu, Li, and Xiao]{chen2024talk}
Yuyan Chen, Yichen Yuan, Panjun Liu, Dayiheng Liu, Qinghao Guan, Mengfei Guo, Haiming Peng, Bang Liu, Zhixu Li, and Yanghua Xiao.
\newblock Talk funny! a large-scale humor response dataset with chain-of-humor interpretation.
\newblock In \emph{Proceedings of the AAAI Conference on Artificial Intelligence}, volume~38, pages 17826--17834, 2024{\natexlab{a}}.

\bibitem[Chen et~al.(2024{\natexlab{b}})Chen, Deng, Yuan, Ji, and Gu]{chen2024self}
Zixiang Chen, Yihe Deng, Huizhuo Yuan, Kaixuan Ji, and Quanquan Gu.
\newblock Self-play fine-tuning converts weak language models to strong language models.
\newblock \emph{arXiv preprint arXiv:2401.01335}, 2024{\natexlab{b}}.

\bibitem[Christiano et~al.(2017)Christiano, Leike, Brown, Martic, Legg, and Amodei]{christiano2017deep}
Paul~F Christiano, Jan Leike, Tom Brown, Miljan Martic, Shane Legg, and Dario Amodei.
\newblock Deep reinforcement learning from human preferences.
\newblock \emph{Advances in neural information processing systems}, 30, 2017.

\bibitem[De~Marez et~al.(2024)De~Marez, Winters, and Terryn]{de2024thinc}
Victor De~Marez, Thomas Winters, and Ayla~Rigouts Terryn.
\newblock Thinc: A theory-driven framework for computational humor detection.
\newblock \emph{arXiv preprint arXiv:2409.01232}, 2024.

\bibitem[Dumoulin et~al.(2023)Dumoulin, Johnson, Castro, Larochelle, and Dauphin]{dumoulin2023density}
Vincent Dumoulin, Daniel~D Johnson, Pablo~Samuel Castro, Hugo Larochelle, and Yann Dauphin.
\newblock A density estimation perspective on learning from pairwise human preferences.
\newblock \emph{arXiv preprint arXiv:2311.14115}, 2023.

\bibitem[Ethayarajh et~al.(2022)Ethayarajh, Choi, and Swayamdipta]{ethayarajh2022understanding}
Kawin Ethayarajh, Yejin Choi, and Swabha Swayamdipta.
\newblock Understanding dataset difficulty with $\mathcal{V}$-usable information.
\newblock In \emph{International Conference on Machine Learning}, pages 5988--6008. PMLR, 2022.

\bibitem[Ethayarajh et~al.(2024)Ethayarajh, Xu, Muennighoff, Jurafsky, and Kiela]{ethayarajh2024kto}
Kawin Ethayarajh, Winnie Xu, Niklas Muennighoff, Dan Jurafsky, and Douwe Kiela.
\newblock Kto: Model alignment as prospect theoretic optimization.
\newblock \emph{arXiv preprint arXiv:2402.01306}, 2024.

\bibitem[Gao et~al.(2023)Gao, Schulman, and Hilton]{gao2023scaling}
Leo Gao, John Schulman, and Jacob Hilton.
\newblock Scaling laws for reward model overoptimization.
\newblock In \emph{International Conference on Machine Learning}, pages 10835--10866. PMLR, 2023.

\bibitem[Gulcehre et~al.(2023)Gulcehre, Paine, Srinivasan, Konyushkova, Weerts, Sharma, Siddhant, Ahern, Wang, Gu, et~al.]{gulcehre2023reinforced}
Caglar Gulcehre, Tom~Le Paine, Srivatsan Srinivasan, Ksenia Konyushkova, Lotte Weerts, Abhishek Sharma, Aditya Siddhant, Alex Ahern, Miaosen Wang, Chenjie Gu, et~al.
\newblock Reinforced self-training (rest) for language modeling.
\newblock \emph{arXiv preprint arXiv:2308.08998}, 2023.

\bibitem[Hejna et~al.(2023)Hejna, Rafailov, Sikchi, Finn, Niekum, Knox, and Sadigh]{hejna2023contrastive}
Joey Hejna, Rafael Rafailov, Harshit Sikchi, Chelsea Finn, Scott Niekum, W~Bradley Knox, and Dorsa Sadigh.
\newblock Contrastive prefence learning: Learning from human feedback without rl.
\newblock \emph{arXiv preprint arXiv:2310.13639}, 2023.

\bibitem[Hessel et~al.(2022)Hessel, Marasovi{\'c}, Hwang, Lee, Da, Zellers, Mankoff, and Choi]{hessel2022androids}
Jack Hessel, Ana Marasovi{\'c}, Jena~D Hwang, Lillian Lee, Jeff Da, Rowan Zellers, Robert Mankoff, and Yejin Choi.
\newblock Do androids laugh at electric sheep? humor" understanding" benchmarks from the new yorker caption contest.
\newblock \emph{arXiv preprint arXiv:2209.06293}, 2022.

\bibitem[Hu et~al.(2021)Hu, Shen, Wallis, Allen-Zhu, Li, Wang, Wang, and Chen]{hu2021lora}
Edward~J Hu, Yelong Shen, Phillip Wallis, Zeyuan Allen-Zhu, Yuanzhi Li, Shean Wang, Lu~Wang, and Weizhu Chen.
\newblock Lora: Low-rank adaptation of large language models.
\newblock \emph{arXiv preprint arXiv:2106.09685}, 2021.

\bibitem[Hu et~al.(2022)Hu, Ying, Wang, and Lyu]{hu2022sum}
Shu Hu, Yiming Ying, Xin Wang, and Siwei Lyu.
\newblock Sum of ranked range loss for supervised learning.
\newblock \emph{Journal of Machine Learning Research}, 23\penalty0 (112):\penalty0 1--44, 2022.

\bibitem[Jamieson et~al.(2015)Jamieson, Jain, Fernandez, Glattard, and Nowak]{jamieson2015next}
Kevin~G Jamieson, Lalit Jain, Chris Fernandez, Nicholas~J Glattard, and Rob Nowak.
\newblock Next: A system for real-world development, evaluation, and application of active learning.
\newblock In \emph{Advances in Neural Information Processing Systems}, pages 2656--2664, 2015.

\bibitem[Jentzsch and Kersting(2023)]{jentzsch2023chatgpt}
Sophie Jentzsch and Kristian Kersting.
\newblock Chatgpt is fun, but it is not funny! humor is still challenging large language models.
\newblock \emph{arXiv preprint arXiv:2306.04563}, 2023.

\bibitem[Jiang et~al.(2023)Jiang, Sablayrolles, Mensch, Bamford, Chaplot, Casas, Bressand, Lengyel, Lample, Saulnier, et~al.]{jiang2023mistral}
Albert~Q Jiang, Alexandre Sablayrolles, Arthur Mensch, Chris Bamford, Devendra~Singh Chaplot, Diego de~las Casas, Florian Bressand, Gianna Lengyel, Guillaume Lample, Lucile Saulnier, et~al.
\newblock Mistral 7b.
\newblock \emph{arXiv preprint arXiv:2310.06825}, 2023.

\bibitem[King et~al.(2013)King, Jha, Radev, and Mankoff]{king2013random}
Ben King, Rahul Jha, Dragomir Radev, and Robert Mankoff.
\newblock Random walk factoid annotation for collective discourse.
\newblock In \emph{Proceedings of the 51st Annual Meeting of the Association for Computational Linguistics (Volume 2: Short Papers)}, pages 249--254, 2013.

\bibitem[Kirk et~al.(2023)Kirk, Mediratta, Nalmpantis, Luketina, Hambro, Grefenstette, and Raileanu]{kirk2023understanding}
Robert Kirk, Ishita Mediratta, Christoforos Nalmpantis, Jelena Luketina, Eric Hambro, Edward Grefenstette, and Roberta Raileanu.
\newblock Understanding the effects of rlhf on llm generalisation and diversity.
\newblock \emph{arXiv preprint arXiv:2310.06452}, 2023.

\bibitem[Lee et~al.(2023)Lee, Phatale, Mansoor, Lu, Mesnard, Bishop, Carbune, and Rastogi]{lee2023rlaif}
Harrison Lee, Samrat Phatale, Hassan Mansoor, Kellie Lu, Thomas Mesnard, Colton Bishop, Victor Carbune, and Abhinav Rastogi.
\newblock Rlaif: Scaling reinforcement learning from human feedback with ai feedback.
\newblock \emph{arXiv preprint arXiv:2309.00267}, 2023.

\bibitem[Li et~al.(2015)Li, Galley, Brockett, Gao, and Dolan]{li2015diversity}
Jiwei Li, Michel Galley, Chris Brockett, Jianfeng Gao, and Bill Dolan.
\newblock A diversity-promoting objective function for neural conversation models.
\newblock \emph{arXiv preprint arXiv:1510.03055}, 2015.

\bibitem[Liu et~al.(2024)Liu, Li, Wu, and Lee]{liu2024visual}
Haotian Liu, Chunyuan Li, Qingyang Wu, and Yong~Jae Lee.
\newblock Visual instruction tuning.
\newblock \emph{Advances in neural information processing systems}, 36, 2024.

\bibitem[Liu et~al.(2023)Liu, Zhao, Joshi, Khalman, Saleh, Liu, and Liu]{liu2023statistical}
Tianqi Liu, Yao Zhao, Rishabh Joshi, Misha Khalman, Mohammad Saleh, Peter~J Liu, and Jialu Liu.
\newblock Statistical rejection sampling improves preference optimization.
\newblock \emph{arXiv preprint arXiv:2309.06657}, 2023.

\bibitem[Mason et~al.(2020)Mason, Jain, Tripathy, and Nowak]{mason2020finding}
Blake Mason, Lalit Jain, Ardhendu Tripathy, and Robert Nowak.
\newblock Finding all $\epsilon$-good arms in stochastic bandits.
\newblock \emph{Advances in Neural Information Processing Systems}, 33:\penalty0 20707--20718, 2020.

\bibitem[Mukobi et~al.(2023)Mukobi, Chatain, Fong, Windesheim, Kutyniok, Bhatia, and Alberti]{mukobi2023superhf}
Gabriel Mukobi, Peter Chatain, Su~Fong, Robert Windesheim, Gitta Kutyniok, Kush Bhatia, and Silas Alberti.
\newblock Superhf: Supervised iterative learning from human feedback.
\newblock \emph{arXiv preprint arXiv:2310.16763}, 2023.

\bibitem[Munos et~al.(2023)Munos, Valko, Calandriello, Azar, Rowland, Guo, Tang, Geist, Mesnard, Michi, et~al.]{munos2023nash}
R{\'e}mi Munos, Michal Valko, Daniele Calandriello, Mohammad~Gheshlaghi Azar, Mark Rowland, Zhaohan~Daniel Guo, Yunhao Tang, Matthieu Geist, Thomas Mesnard, Andrea Michi, et~al.
\newblock Nash learning from human feedback.
\newblock \emph{arXiv preprint arXiv:2312.00886}, 2023.

\bibitem[Nakano et~al.(2021)Nakano, Hilton, Balaji, Wu, Ouyang, Kim, Hesse, Jain, Kosaraju, Saunders, et~al.]{nakano2021webgpt}
Reiichiro Nakano, Jacob Hilton, Suchir Balaji, Jeff Wu, Long Ouyang, Christina Kim, Christopher Hesse, Shantanu Jain, Vineet Kosaraju, William Saunders, et~al.
\newblock Webgpt: Browser-assisted question-answering with human feedback.
\newblock \emph{arXiv preprint arXiv:2112.09332}, 2021.

\bibitem[Ouyang et~al.(2022)Ouyang, Wu, Jiang, Almeida, Wainwright, Mishkin, Zhang, Agarwal, Slama, Ray, et~al.]{ouyang2022training}
Long Ouyang, Jeffrey Wu, Xu~Jiang, Diogo Almeida, Carroll Wainwright, Pamela Mishkin, Chong Zhang, Sandhini Agarwal, Katarina Slama, Alex Ray, et~al.
\newblock Training language models to follow instructions with human feedback.
\newblock \emph{Advances in neural information processing systems}, 35:\penalty0 27730--27744, 2022.

\bibitem[Palan and Schitter(2018)]{palan2018prolific}
Stefan Palan and Christian Schitter.
\newblock Prolific. ac—a subject pool for online experiments.
\newblock \emph{Journal of Behavioral and Experimental Finance}, 17:\penalty0 22--27, 2018.

\bibitem[Radev et~al.(2015)Radev, Stent, Tetreault, Pappu, Iliakopoulou, Chanfreau, de~Juan, Vallmitjana, Jaimes, Jha, et~al.]{radev2015humor}
Dragomir Radev, Amanda Stent, Joel Tetreault, Aasish Pappu, Aikaterini Iliakopoulou, Agustin Chanfreau, Paloma de~Juan, Jordi Vallmitjana, Alejandro Jaimes, Rahul Jha, et~al.
\newblock Humor in collective discourse: Unsupervised funniness detection in the new yorker cartoon caption contest.
\newblock \emph{arXiv preprint arXiv:1506.08126}, 2015.

\bibitem[Rafailov et~al.(2024)Rafailov, Sharma, Mitchell, Manning, Ermon, and Finn]{rafailov2024direct}
Rafael Rafailov, Archit Sharma, Eric Mitchell, Christopher~D Manning, Stefano Ermon, and Chelsea Finn.
\newblock Direct preference optimization: Your language model is secretly a reward model.
\newblock \emph{Advances in Neural Information Processing Systems}, 36, 2024.

\bibitem[Reimers and Gurevych(2019)]{reimers2019sentence}
Nils Reimers and Iryna Gurevych.
\newblock Sentence-bert: Sentence embeddings using siamese bert-networks.
\newblock \emph{arXiv preprint arXiv:1908.10084}, 2019.

\bibitem[Rosset et~al.(2024)Rosset, Cheng, Mitra, Santacroce, Awadallah, and Xie]{rosset2024direct}
Corby Rosset, Ching-An Cheng, Arindam Mitra, Michael Santacroce, Ahmed Awadallah, and Tengyang Xie.
\newblock Direct nash optimization: Teaching language models to self-improve with general preferences.
\newblock \emph{arXiv preprint arXiv:2404.03715}, 2024.

\bibitem[Schulman et~al.(2017)Schulman, Wolski, Dhariwal, Radford, and Klimov]{schulman2017proximal}
John Schulman, Filip Wolski, Prafulla Dhariwal, Alec Radford, and Oleg Klimov.
\newblock Proximal policy optimization algorithms.
\newblock \emph{arXiv preprint arXiv:1707.06347}, 2017.

\bibitem[Shahaf et~al.(2015)Shahaf, Horvitz, and Mankoff]{shahaf2015inside}
Dafna Shahaf, Eric Horvitz, and Robert Mankoff.
\newblock Inside jokes: Identifying humorous cartoon captions.
\newblock In \emph{Proceedings of the 21th ACM SIGKDD international conference on knowledge discovery and data mining}, pages 1065--1074, 2015.

\bibitem[Sharma et~al.(2024)Sharma, Keh, Mitchell, Finn, Arora, and Kollar]{sharma2024critical}
Archit Sharma, Sedrick Keh, Eric Mitchell, Chelsea Finn, Kushal Arora, and Thomas Kollar.
\newblock A critical evaluation of ai feedback for aligning large language models.
\newblock \emph{arXiv preprint arXiv:2402.12366}, 2024.

\bibitem[Sievert et~al.(2017)Sievert, Ross, Jain, Jamieson, Nowak, and Mankoff]{sievert2017next}
Scott Sievert, Daniel Ross, Lalit Jain, Kevin Jamieson, Robert Nowak, and Robert Mankoff.
\newblock Next: A system to easily connect crowdsourcing and adaptive data collection.
\newblock In \emph{SciPy}, pages 113--119, 2017.

\bibitem[Siththaranjan et~al.(2023)Siththaranjan, Laidlaw, and Hadfield-Menell]{siththaranjan2023distributional}
Anand Siththaranjan, Cassidy Laidlaw, and Dylan Hadfield-Menell.
\newblock Distributional preference learning: Understanding and accounting for hidden context in rlhf.
\newblock \emph{arXiv preprint arXiv:2312.08358}, 2023.

\bibitem[Song et~al.(2024)Song, Yu, Li, Yu, Huang, Li, and Wang]{song2024preference}
Feifan Song, Bowen Yu, Minghao Li, Haiyang Yu, Fei Huang, Yongbin Li, and Houfeng Wang.
\newblock Preference ranking optimization for human alignment.
\newblock In \emph{Proceedings of the AAAI Conference on Artificial Intelligence}, volume~38, pages 18990--18998, 2024.

\bibitem[Stasaski and Hearst(2022)]{stasaski2022semantic}
Katherine Stasaski and Marti~A Hearst.
\newblock Semantic diversity in dialogue with natural language inference.
\newblock \emph{arXiv preprint arXiv:2205.01497}, 2022.

\bibitem[Swamy et~al.(2024)Swamy, Dann, Kidambi, Wu, and Agarwal]{swamy2024minimaximalist}
Gokul Swamy, Christoph Dann, Rahul Kidambi, Zhiwei~Steven Wu, and Alekh Agarwal.
\newblock A minimaximalist approach to reinforcement learning from human feedback.
\newblock \emph{arXiv preprint arXiv:2401.04056}, 2024.

\bibitem[Tanczos et~al.(2017)Tanczos, Nowak, and Mankoff]{tanczos2017kl}
Ervin Tanczos, Robert Nowak, and Bob Mankoff.
\newblock A kl-lucb bandit algorithm for large-scale crowdsourcing.
\newblock In \emph{Proceedings of the 31st International Conference on Neural Information Processing Systems}, pages 5896--5905, 2017.

\bibitem[Tang et~al.(2024)Tang, Guo, Zheng, Calandriello, Munos, Rowland, Richemond, Valko, Pires, and Piot]{tang2024generalized}
Yunhao Tang, Zhaohan~Daniel Guo, Zeyu Zheng, Daniele Calandriello, R{\'e}mi Munos, Mark Rowland, Pierre~Harvey Richemond, Michal Valko, Bernardo~{\'A}vila Pires, and Bilal Piot.
\newblock Generalized preference optimization: A unified approach to offline alignment.
\newblock \emph{arXiv preprint arXiv:2402.05749}, 2024.

\bibitem[Touvron et~al.(2023)Touvron, Martin, Stone, Albert, Almahairi, Babaei, Bashlykov, Batra, Bhargava, Bhosale, et~al.]{touvron2023llama}
Hugo Touvron, Louis Martin, Kevin Stone, Peter Albert, Amjad Almahairi, Yasmine Babaei, Nikolay Bashlykov, Soumya Batra, Prajjwal Bhargava, Shruti Bhosale, et~al.
\newblock Llama 2: Open foundation and fine-tuned chat models.
\newblock \emph{arXiv preprint arXiv:2307.09288}, 2023.

\bibitem[von Werra et~al.()von Werra, Belkada, Tunstall, Beeching, Thrush, and Lambert]{von_Werra_TRL_Transformer_Reinforcement}
Leandro von Werra, Younes Belkada, Lewis Tunstall, Edward Beeching, Tristan Thrush, and Nathan Lambert.
\newblock {TRL: Transformer Reinforcement Learning}.
\newblock URL \url{https://github.com/huggingface/trl}.

\bibitem[Wu et~al.(2024)Wu, Hu, Shi, Dziri, Suhr, Ammanabrolu, Smith, Ostendorf, and Hajishirzi]{wu2024fine}
Zeqiu Wu, Yushi Hu, Weijia Shi, Nouha Dziri, Alane Suhr, Prithviraj Ammanabrolu, Noah~A Smith, Mari Ostendorf, and Hannaneh Hajishirzi.
\newblock Fine-grained human feedback gives better rewards for language model training.
\newblock \emph{Advances in Neural Information Processing Systems}, 36, 2024.

\bibitem[Xu et~al.(2023)Xu, Rosset, Del~Corro, Mahajan, McAuley, Neville, Awadallah, and Rao]{xu2023contrastive}
Canwen Xu, Corby Rosset, Luciano Del~Corro, Shweti Mahajan, Julian McAuley, Jennifer Neville, Ahmed~Hassan Awadallah, and Nikhil Rao.
\newblock Contrastive post-training large language models on data curriculum.
\newblock \emph{arXiv preprint arXiv:2310.02263}, 2023.

\bibitem[Xu et~al.(2024)Xu, Yuan, Chen, and Yang]{xu2024good}
Zhijun Xu, Siyu Yuan, Lingjie Chen, and Deqing Yang.
\newblock " a good pun is its own reword": Can large language models understand puns?
\newblock \emph{arXiv preprint arXiv:2404.13599}, 2024.

\bibitem[Yang et~al.(2017)Yang, Ramdas, Jamieson, and Wainwright]{yang2017framework}
Fanny Yang, Aaditya Ramdas, Kevin~G Jamieson, and Martin~J Wainwright.
\newblock A framework for multi-a (rmed)/b (andit) testing with online fdr control.
\newblock \emph{Advances in Neural Information Processing Systems}, 30, 2017.

\bibitem[Yang et~al.(2023)Yang, Klein, Celikyilmaz, Peng, and Tian]{yang2023rlcd}
Kevin Yang, Dan Klein, Asli Celikyilmaz, Nanyun Peng, and Yuandong Tian.
\newblock Rlcd: Reinforcement learning from contrast distillation for language model alignment.
\newblock \emph{arXiv preprint arXiv:2307.12950}, 2023.

\bibitem[Yin et~al.(2024)Yin, Wang, Gu, Huang, Chen, and Zhou]{yin2024relative}
Yueqin Yin, Zhendong Wang, Yi~Gu, Hai Huang, Weizhu Chen, and Mingyuan Zhou.
\newblock Relative preference optimization: Enhancing llm alignment through contrasting responses across identical and diverse prompts.
\newblock \emph{arXiv preprint arXiv:2402.10958}, 2024.

\bibitem[Yuan et~al.(2024)Yuan, Pang, Cho, Sukhbaatar, Xu, and Weston]{yuan2024self}
Weizhe Yuan, Richard~Yuanzhe Pang, Kyunghyun Cho, Sainbayar Sukhbaatar, Jing Xu, and Jason Weston.
\newblock Self-rewarding language models.
\newblock \emph{arXiv preprint arXiv:2401.10020}, 2024.

\bibitem[Yuan et~al.(2023)Yuan, Yuan, Tan, Wang, Huang, and Huang]{yuan2023rrhf}
Zheng Yuan, Hongyi Yuan, Chuanqi Tan, Wei Wang, Songfang Huang, and Fei Huang.
\newblock Rrhf: Rank responses to align language models with human feedback without tears.
\newblock \emph{arXiv preprint arXiv:2304.05302}, 2023.

\bibitem[Zhao et~al.(2023)Zhao, Joshi, Liu, Khalman, Saleh, and Liu]{zhao2023slic}
Yao Zhao, Rishabh Joshi, Tianqi Liu, Misha Khalman, Mohammad Saleh, and Peter~J Liu.
\newblock Slic-hf: Sequence likelihood calibration with human feedback.
\newblock \emph{arXiv preprint arXiv:2305.10425}, 2023.

\bibitem[Zhong et~al.(2024)Zhong, Huang, Gao, Wen, Lin, Zitnik, and Zhou]{zhong2024let}
Shanshan Zhong, Zhongzhan Huang, Shanghua Gao, Wushao Wen, Liang Lin, Marinka Zitnik, and Pan Zhou.
\newblock Let's think outside the box: Exploring leap-of-thought in large language models with creative humor generation.
\newblock In \emph{Proceedings of the IEEE/CVF Conference on Computer Vision and Pattern Recognition}, pages 13246--13257, 2024.

\bibitem[Zhou et~al.(2024)Zhou, Liu, Xu, Iyer, Sun, Mao, Ma, Efrat, Yu, Yu, et~al.]{zhou2024lima}
Chunting Zhou, Pengfei Liu, Puxin Xu, Srinivasan Iyer, Jiao Sun, Yuning Mao, Xuezhe Ma, Avia Efrat, Ping Yu, Lili Yu, et~al.
\newblock Lima: Less is more for alignment.
\newblock \emph{Advances in Neural Information Processing Systems}, 36, 2024.

\bibitem[Zhu et~al.(2023)Zhu, Frick, Wu, Zhu, and Jiao]{starling2023}
Banghua Zhu, Evan Frick, Tianhao Wu, Hanlin Zhu, and Jiantao Jiao.
\newblock Starling-7b: Improving llm helpfulness \& harmlessness with rlaif, November 2023.

\bibitem[Ziegler et~al.(2019)Ziegler, Stiennon, Wu, Brown, Radford, Amodei, Christiano, and Irving]{ziegler2019fine}
Daniel~M Ziegler, Nisan Stiennon, Jeffrey Wu, Tom~B Brown, Alec Radford, Dario Amodei, Paul Christiano, and Geoffrey Irving.
\newblock Fine-tuning language models from human preferences.
\newblock \emph{arXiv preprint arXiv:1909.08593}, 2019.

\end{thebibliography}
\bibliographystyle{plainnat}

\newpage

\appendix

\section{Links to Resources} \label{apx:links}
Our dataset is available at \url{https://huggingface.co/datasets/yguooo/newyorker_caption_ranking} under Creative Commons Attribution Non Commercial 4.0. Our codebase is available at \url{https://github.com/yguooo/cartoon-caption-generation} under Apache 2.0.

\section{Language Model Prompts} \label{apx:prompts}
\subsection{Description Generation}
\label{apx:description_generation}
We use GPT-4o to generate descriptions for each cartoon. In the dataset from \citet{hessel2022androids} each cartoon has a canny description, an uncanny description, a location, and a list of entity. Entity are words that is related to the cartoon. We used the five shot method to generate a set of descriptions. The five examples are randomly selected from the testing set, and we use the these same five example for every cartoon descriptions generation. An example of our prompt is shown below. 

\begin{center}
\fbox{
\begin{minipage}{.95\linewidth}
\textbf{User}: In this task, you will see a cartoon, then write two descriptions about the cartoon, one uncanny description and one canny description, then write the cartoon's location, and the entities of the cartoon. I am going to give you five examples first and you write the last sets of description.

\textbf{User}: $<$Insert Cartoon Image$>$

\textbf{Assistant}: The canny description is $<$insert canny description$>$ and the uncanny description is $<$insert uncanny description$>$, and the cartoon's location is $<$insert location$>$, and the entities of the cartoon are $<$insert entities$>$

......\textit{Repeat user/assistant for four more examples}......

\textbf{User}: $<$Insert Cartoon Image$>$. The set of description is
\end{minipage}
}
\end{center}

\begin{table}[h]
    \centering
    
    \caption{Examples of Generated Cartoon Descriptions}
    \begin{tabularx}{\textwidth}{lXX}
        \toprule\\
        Type of descriptions  & GPT-4o & Human Written~\citep{hessel2022androids}\\
        \midrule
        Canny description       & \texttt{A knight in armor is riding a horse, holding a lance with a traffic light on top. A line of businessmen in suits follows behind him.} & \texttt{There are two men on a horse. They are wearing soldier outfits. Businessmen follow behind them.}\\
        Uncanny Description     & \texttt{It's unusual to see a medieval knight leading modern businessmen as if going into battle.} & \texttt{There are businessmen following a two guys on horses who are soldiers.}\\
        Location              & \texttt{an open field} & \texttt{a hilly path}\\
        Entities              &\texttt{Knight, Horse, Businessmen, Traffic light} & \texttt{Warrior, Horses in warfare, Businessperson}\\
        \bottomrule
    \end{tabularx}

    \label{tab:description}
\end{table}

\subsection{Caption Evaluation}
 We evaluate various models that generate captions by comparing the generated captions against four groups of human contestant entries at different ranking levels, which include top10, \#200-\#209, \#1000-\#1009, and contestant median. As concluded based on Table~\ref{tab:ranking}, we use GPT4-Turbo as evaluator with descriptions from \citet{hessel2022androids} in Overall Comparison and GPT4o-vision as evaluator with raw cartoon images in Best Pick Comparison. For both group comparison methods, we utilize the 5-shot in-context prompting technique, as mentioned in Section~\ref{ssec:eval}. 
 
 An example of Overall Comparison is shown below.

\begin{center}
\fbox{
\begin{minipage}{.95\linewidth}
\textbf{System}: You are a judge for the new yorker cartoon caption contest.

\textbf{User}: In this task, you will see two description for a cartoon. Then, you will see two captions that were written about the cartoon. Then you will choose which captions is funnier. I am going to give you five examples first and you answer the last example with either A or B.

\textbf{User}: For example, the descriptions for the images are $<$Insert Canny Description$>$ and $<$Insert Uncanny Description$>$.  The two captions are A: $<$Insert CaptionA$>$ B: $<$Insert CaptionB$>$

\textbf{Assistant}: The caption that is funnier is $<$Insert Answer$>$

......\textit{Repeat user/assistant for four more examples}......

\textbf{User}: The descriptions for the images are $<$Insert Canny Description$>$ and $<$Insert Uncanny Description$>$. The two groups of captions are group A: $<$Insert Caption Group A$>$ group B: $<$Insert Caption Group B$>$ 

\textbf{User}: Choose the group of captions that is funnier. Answer with only one letter A or B, and nothing else.

\end{minipage}
}
\end{center}

An example of Best Pick Comparison is shown below.

\begin{center}
\fbox{
\begin{minipage}{.95\linewidth}
\textbf{System}: You are a judge for the new yorker cartoon caption contest. Your job is to find the funniest caption.

\textbf{User}: In this task, you will see a cartoon first and two captions that were written about it then. The task is to choose which caption is funnier. I am going to show you five cartoons, corresponding captions and their answers first. In the end, for the last cartoon, answer with only one letter A or B, and nothing else.

\textbf{User}: $<$Insert Cartoon Image$>$

\textbf{User}: For this example, the two captions are A: $<$Insert CaptionA$>$ B: $<$Insert CaptionB$>$. The answer is

\textbf{Assistant}: $<$Insert Answer$>$

......\textit{Repeat user/assistant for four more examples}......

\textbf{User}: $<$Insert Cartoon Image$>$ 

\textbf{User}: Find the funniest caption for each group. Then choose the funnier group based on these funniest captions. Think step by step but finish the last line of your answer with only one letter A or B, and nothing else. A: $<$Insert Caption A$>$ or B: $<$Insert Caption B$>$

\end{minipage}
}
\end{center}

\subsection{Caption Generation}
We used GPT-3.5-turbo, Claude-3-opus, and GPT-4-o to generate captions for each cartoons. We first use the system role to prompt it to generate 10 captions. Then we provide the image descriptions and then the image itself. For GPT-3.5-turbo, we simply only provided the image descriptions. For GPT-4-o, we have two versions where in one we provide the image itself, and the other we only provided the image descriptions. For Claude, we always provide both image description and image itself.

\begin{center}
\fbox{
\begin{minipage}{.95\linewidth}
\textbf{System}: I want you to act as a sophisticated reader of The New Yorker Magazine. You are competing in The New Yorker Cartoon Caption Contest.  Your task is to generate funny captions for a cartoon. Here are some ideas for developing funny captions.   First think about characteristics associated with the objects and people featured in the cartoon. Then consider what are the unusual or absurd elements in the cartoon. It might help to imagine conversations between the characters.  Then think about funny and non-obvious connections that can be made between the objects and characters. Try to come up with funny captions that fit the cartoon, but are not too direct.  It may be funnier if the person reading the caption has to think a little bit to get the joke. Next, I will describe a cartoon image and then you should generate 10 funny captions for the cartoon along with an explanation for each.

\textbf{User}: $<$Insert Cartoon Image$>$

\textbf{User}: The cartoon's description is: $<$insert canny description$>$.The uncanny description is: $<$insert uncanny description$>$. The location of the cartoon is:$<$insert location$>$. The entities of the cartoon are: $<$insert image entities$>$

\end{minipage}
}
\end{center}

\section{Additional Experiment Setups} \label{apx:additional_setup}

\subsection{Human Experiment Details} \label{apx:human}

Each participant provided informed consent in compliance with our
Institutional IRB and was compensated for their
time. We paid participants \$12 an hour and spent about \$600 on data collection. The following instructions were used for the human experiments.

\subsubsection{Human Pairwise with description generated by GPT4o-vision}

In each trial of this task, you will see a description of a cartoon and two captions: the cartoon description is on the top, and the two caption choices are beneath the cartoon description. For each trial, please select the caption that is the funniest for the cartoon.

\subsubsection{Human Pairwise with Cartoon Image}
In each trial of this task, you will see one cartoon and two captions: the cartoon is on top, and the two caption choices are beneath the cartoon. For each trial, please select the caption that is the funniest for the cartoon.

\subsubsection{Human Group (Overall) with description generated by GPT4o-vision}
In each trial of this task, you will see a description of a cartoon and two groups of captions: the cartoon description is on the top, and the two grouped caption choices are beneath the cartoon description. For each trial, please select the group of captions that is the funniest for the cartoon.

\subsubsection{Human Group (Overall) with Cartoon Image}

In each trial of this task, you will see a cartoon and two groups of captions: the cartoon is on the top, and the two grouped caption choices are beneath the cartoon. For each trial, please select the group of captions that is the funniest for the cartoon.

\subsubsection{Human Group (Best Pick) with description generated by GPT4o-vision}
In each trial of this task, you will see a description of a cartoon and two groups of captions: the cartoon description is on the top, and the two grouped caption choices are beneath the cartoon description. For each trial, please select the group of captions that contains the funniest caption for the cartoon. First, pick the funniest caption in each group, and then compare between the two captions to pick the funniest group.

\subsubsection{Human Group (Best Pick) with Cartoon Image}

In each trial of this task, you will see a cartoon and two groups of captions: the cartoon is on the top, and the two grouped caption choices are beneath the cartoon. For each trial, please select the group of captions that contains the funniest caption for the cartoon. First, pick the funniest caption in each group, and then compare between the two captions to pick the funniest group.

\subsubsection{Human top 10 vs Claude generated captions}
In each trial of this task, you will see a cartoon and two groups of captions: the cartoon is on the top, and the two grouped caption choices are beneath the cartoon. For each trial, <strong>please select the group of captions that is the funniest for the cartoon.

\subsection{Recalibration of GPT Models for Ranking}
For group comparisons without chain of thought, we observe a strong bias of GPT4 models choosing $A$ over $B$. In other words, for some examples, the model always chooses option $A$ even after we flip the two groups. Therefore, this suggests we need to calibrate the model predictions. We adopt a simple approach by readjusting the decision threshold. Let $s_i^A, s_i^B$ denote the log probabilities of choosing $A$ and $B$ by the GPT4 model for two groups of human submitted captions $x_i^A$ and $x_i^B$ respectively. We use a small validation set of $m$ examples $\{x_i^A, x_i^B\}_{i=1}^m$ with sigmoid scores $\{s_i^A, s_i^B\}_{i=1}^m$ and ground truth preference by the crowd denoted as $\{y_i\in \{A, B\}\}_{i=1}^m$. The current decision rule takes the form of $\widehat{f}(x_i^A, x_i^B) = \begin{cases}
    A & \text{if } s_i^A - s_i^B > 0 \\
    B & \text{otherwise}
\end{cases}$.

We simply set a different threshold $\tau$, which induces $\widehat{f}_\tau(x_i^A, x_i^B) = \begin{cases}
    A & \text{if } s_i^A - s_i^B > \tau \\
    B & \text{otherwise}
\end{cases}$. The threshold $\tau^*$ is chosen so that the accuracy over the validation set is maximized:
\begin{align*}
    \tau^* = \argmax_\tau \sum_{i=1}^m \1\{y_i = \widehat{f}_\tau(x_i^A, x_i^B)\}.
\end{align*}
Ties are broken arbitrarily above. We then use the recalibrated decision rule with $\tau^*$ for all of our evaluations.

\subsection{Finetuning Experiment Details} \label{apx:exp_details}

Our training and test split for finetuning range from contest 530 to 890. In particular, our dataset includes all the data of \citep{hessel2022androids} with ranking information within this range. (\citep{hessel2022androids} only contains contests up to \#763.) Thus, we choose our test split to be the combination of testing (47 contests) and validation split (44 contests) of \citep{hessel2022androids} within the 530-890 range. The rest available contests form our training split. 

Our finetuning methods are trained from Mistral 7B Instruct v0.1 and LLaVa v1.6 Mistral (multimodal case) via LoRA updates~\citep{hu2021lora}. We use a variant of Mistral 7b model as our initial reward model to finetune from \footnote{We use the pretrained reward model from \url{https://huggingface.co/weqweasdas/RM-Mistral-7B}}. The choice of reward is based on our benchmarking results of top reward models on our caption generation dataset (\cref{tab:reward_model_benchmark}). For SFT methods, we train on 1000 pairs of captions from each contest, with the preferred caption from the top 1000 captions and the alternative randomly sampled from the rest. For reward modeling, DPO and RLHF, we train on 1000 pairs of captions with three standard deviations apart according to~\cref{eq:3_std} per contest. Additionally, we train our model using the default choice of optimizer from TRL up to 1 epoch. Then, we search for the best hyper-parameter over the neighborhood of default parameters and pick the best performing model under our GPT-based group comparison metrics. For our reward model, we pick the best model based on the reward evaluation on the holdout set. For both pretrained and finetuned models, we use the same generation configuration file with temperature 0.7, top-p sampling probability 0.95, repetition penalty 1.15. When evaluating using the Best-of-N (BoN) method, we pick the top 10 captions based on the trained reward model, out of 50 generated candidates from caption generation models. Our choice of batch is 64 for SFT and reward model, and 128 for all other settings.  

During the training process of DPO, PPO, SFT, we create a separate padding tokens and resize the token embedding of the pretrained model so that the text generation can terminate properly. Furthermore, in the loss design of SFT case, we only evaluate the next-token prediction loss on the caption segment, as all the training texts contain similar prompts. Since we only reported the iteration with the best results, early stopping occurs before a single epoch for the choice of best iterations.

We also noted that PPO performs the best when starting from the pretrained Mistral Instruct 7B model, whereas DPO performs the best from a sft checkpoint of Mistral. This SFT checkpoint needs to be tuned on simple prompts and does not render a better performance than the sft tuned on the best prompt (with all those descriptions).

\paragraph{Choice of Prompts}
In~\cref{tab:best_prompts}, we document the best prompt we found for each training algorithm. Generally speaking, the zero-shot, SFT, preference learning algorithm each require simpler prompts than the one preceding them. 

\paragraph{Computation Cost}
Finetuning a SFT, DPO, PPO model usually takes 2-4 days to train till convergence on a A100 machine. Evaluating a single number of each scenario cost roughly \$5 on the openai platform.

\section{Crowdsourced Caption Contest Ratings}\label{apx:caption_contest}
%\todo{Lalit}

\begin{algorithm}
\caption{Upper Confidence Bound (UCB) Algorithm}
\begin{algorithmic}[1]\label{alg:ucb}
\STATE \textbf{Initialization:} For each caption $x$, initialize $N_x(0) = 0$ and $\hat{\mu}_x = 0$.
\FOR{$t = 1$ to $T$}
    \STATE Select caption $x_t = \arg\max_{x} \left( \hat{\mu}_x + \sqrt{\frac{2 \ln(4 N_x(t)^2)}{N_x(t)}} \right)$.
    \STATE Observe the reward $r_t\in \{1,2,3\}$ for caption $x_t$.
    \STATE Update the number of times action $x_t$ has been selected: $N_{x_t}(t) = N_{x_t}(t-1) + 1$.
    \STATE Update the empirical mean reward of action $x_t$: 
    \[
    \hat{\mu}_{x_t} = \frac{N_{x_t}(t-1) \cdot \hat{\mu}_{x_t} + r_t}{N_{x_t}(t)}
    \]
\ENDFOR
\end{algorithmic}
\end{algorithm}

As described in the text, we used a UCB~\cite{auer2002using} variant to encourage high-performing captions to recieve the votes. We experimented with standard UCB (see Algorithm~\ref{alg:ucb}) and KL-UCB specifically optimized for discrete rewards~\citep{tanczos2017kl}. The data repository labels datasets according to which algorithm was employed for each contest. In practice, using UCB in high-traffic asynchronous environments faces specific challenges. For example, we wanted to ensure that voters could only vote on one caption at a time, that the model sent batches of captions to users to reduce round trips to the server, and that the underlying model was able to update as frequently as possible. For more details on overcoming such challenges, see ~\citep{jamieson2015next}.

\section{Additional Results}\label{apx:additional_results}
We benchmark the performance of different reward model as in~\cref{tab:reward_model_benchmark}. It is worth noting that our goal here is to understand the effect of the ranking model for the downstream preference learning algorithms, thus we evaluate on the same dataset as in~\cref{eq:3_std} instead of the setting of~\cref{tab:ranking}. weqweasdas/RM-Mistral-7B and Eurus-RM-7B Instruct are the top two models with the highest reward ranking accuracy. We choose to use weqweasdas/RM-Mistral-7B because it generally achieves better ranking accuracy for various data settings that we experimented on. 

In our experiment, we noticed that PPO algorithm requires a much more aggressive early stopping scheme than DPO and SFT. Thus, we further look at the training dynamics of the PPO algorithm in~\cref{tab:ppo_dynamics}. Here, the batch size is 128. It is worth noting that the result at iteration 0 has an lower overall win rate than the zero shot result in~\cref{tab:eval_accs}. The reason is that our PPO and DPO algorithms need to use a simpler prompt as in~\cref{tab:best_prompts} to generate meaningful texts. From~\cref{tab:ppo_dynamics}, we verified the steady increase of the mean reward and decrease of the training loss. However, the improvement on these metrics does not corresponds to an improvement of the overall humorous generation. We hypothesize that this is due to the complex nature of humor and the potential for
out-of-distribution generations when running RLHF. 
\begin{table}[h!]
    \caption{Reward model benchmark}
    \label{tab:reward_model_benchmark}
    \centering
    \begin{tabular}{lc}
    \toprule
          & Reward Ranking Acc (\%)  \\ 
    \midrule
    Mistral-7B Instruct  &   73.17                   \\ 
    Llama-3-8B Instruct&     74.01                 \\ 
    Llama-2-7B Chat  &     72.63                 \\ 
    \midrule
    weqweasdas/RM-Mistral-7B  & \textbf{74.05}                    \\ 
    Eurus-RM-7B  &   \textbf{74.18}                 \\ 
    FsfairX-LLaMA3-RM-v0.1 &  73.72      \\ 
    Qwen1.5-7B-Chat &  72.26    \\ 
    \bottomrule
    \end{tabular}
\end{table}

\begin{table}[h!]
    \caption{Training Dynamics of PPO}
    \label{tab:ppo_dynamics}
    \centering
    \begin{tabular}{lcccccc}
    \toprule
    Iteration      & 0 & 10 & 20 & 30 & 40 & 50  \\ 
    \midrule
    \shortstack[l]{Contestant Median\\(Overall Win Rate (\%)) $\uparrow$}&   17.03 & \textbf{24.73}  & 16.48 & 9.89 & 6.04 & 4.95                   \\
    Mean Reward $\uparrow$ & 0.0057 & 0.0260 & 0.0186 & 0.1309 & 0.1356 & \textbf{0.2587} \\ 
    Loss $\downarrow$  & 0.3592 & 0.2001 & 0.1773 & 0.1709 & 0.0848 & \textbf{0.0584} \\
    \bottomrule
    \end{tabular}
\end{table}

% \begin{table}[]
% \begin{tabular}{|c|cccc|}
% \hline
% & \multicolumn{4}{c|}{BoN Best Pick Win Rate (\%)} \\ \hline
% & \multicolumn{1}{c|}{Top 10} & \multicolumn{1}{c|}{200}   & \multicolumn{1}{c|}{1k}    & Median \\ \hline
% Zero-Shot (Mistral) & \multicolumn{1}{c|}{18.13}  & \multicolumn{1}{c|}{24.73} & \multicolumn{1}{c|}{36.26} & 39.56  \\ \hline
% Finetune (Mistral)  & \multicolumn{1}{c|}{12.09}       & \multicolumn{1}{c|}{18.13}      & \multicolumn{1}{c|}{19.78}      &     32.97   \\ \hline
% DPO (Mistral)       & \multicolumn{1}{c|}{30.77}       & \multicolumn{1}{c|}{35.71}      & \multicolumn{1}{c|}{44.51}      &  47.25      \\ \hline
% PPO (Mistral)       & \multicolumn{1}{c|}{4.94}       & \multicolumn{1}{c|}{9.89}      & \multicolumn{1}{c|}{10.99}      &   16.48     \\ \hline
% LLaVA               & \multicolumn{1}{c|}{}       & \multicolumn{1}{c|}{}      & \multicolumn{1}{c|}{}      &        \\ \hline
% \end{tabular}
% \end{table}

\begin{table}[h!]
\caption{Best choice of prompts for each training algorthm }\label{tab:best_prompts}
\centering
\begin{tabular}{lc}
\toprule
                    & Best Choice of Prompt \\ 
\midrule
Zero-Shot  &    \fbox{
\begin{minipage}{.65\linewidth}
[INST] $<>$ I want you to act as a sophisticated reader of The New Yorker Magazine. You are competing in The New Yorker Cartoon Caption Contest.  Your task is to generate funny captions for a cartoon. Here are some ideas for developing funny captions. \\
First think about characteristics associated with the objects and people featured in the cartoon. Then consider what are the unusual or absurd elements in the cartoon. It might help to imagine conversations between the characters. Then think about funny and non-obvious connections that can be made between the objects and characters. Try to come up with funny captions that fit the cartoon, but are not too direct. It may be funnier if the person reading the caption has to think a little bit to get the joke. Next, I will describe a cartoon image and then you should generate 1 funny caption for the cartoon along with an explanation for each. 

scene: $<$\textit{scene}$>$

description: $<$\textit{description}$>$ 

uncanny description: $<$\textit{uncanny description}$>$

entities: $<$\textit{entities}$>$ $<$$>$ 

funny caption: [/INST] $<$\textit{sample caption}$>$
\end{minipage}
}                   \\ 
\midrule
SFT      &  \fbox{
\begin{minipage}{.65\linewidth}
[INST]I want you to act as a sophisticated reader of The New Yorker Magazine. You are competing in The New Yorker Cartoon Caption Contest.  Your task is to generate funny captions for a cartoon. Here are some ideas for developing funny captions. First think about characteristics associated with the objects and people featured in the cartoon. Then consider what are the unusual or absurd elements in the cartoon. It might help to imagine conversations between the characters.  Then think about funny and non-obvious connections that can be made between the objects and characters. Try to come up with funny captions that fit the cartoon, but are not too direct.  It may be funnier if the person reading the caption has to think a little bit to get the joke. Next, I will describe a cartoon image and then you should generate 1 funny caption for the cartoon[/INST] 

scene: $<$\textit{scene}$>$

description: $<$\textit{description}$>$ 

uncanny description: $<$\textit{uncanny description}$>$

entities: $<$\textit{entities}$>$

funny caption: $<$\textit{sample caption}$>$

\end{minipage}
}     \\ 
\bottomrule
\end{tabular}
\end{table}

\begin{table}[h!]
\centering
\begin{tabular}{lc}
\toprule
                    & Best Choice of Prompt \\ 
\midrule
LLaVA      &
\fbox{
\begin{minipage}{.65\linewidth}
[INST]  I want you to act as a sophisticated reader of The New Yorker Magazine. You are competing in The New Yorker Cartoon Caption Contest.  Your task is to generate funny captions for a cartoon. Here are some ideas for developing funny captions. 

First think about characteristics associated with the objects and people featured in the cartoon. Then consider what are the unusual or absurd elements in the cartoon. It might help to imagine conversations between the characters. Then think about funny and non-obvious connections that can be made between the objects and characters. Try to come up with funny captions that fit the cartoon, but are not too direct. It may be funnier if the person reading the caption has to think a little bit to get the joke. Next, I will provide a cartoon image with descriptions and then you should generate 1 funny caption for the cartoon along with an explanation for each. 

image: $<$\textit{image}$>$

scene: $<$\textit{scene}$>$

description: $<$\textit{description}$>$ 

uncanny description: $<$\textit{uncanny description}$>$

entities: $<$\textit{entities}$>$ 

Generate a funny caption for the image: [/INST] $<$\textit{sample caption}$>$ 
\end{minipage}
}                     \\ 
\midrule
DPO/PPO/Reward Model &  \fbox{
\begin{minipage}{.65\linewidth} 
scene: $<$\textit{scene}$>$

description: $<$\textit{description}$>$ 

uncanny description: $<$\textit{uncanny description}$>$

entities: $<$\textit{entities}$>$ 

funny caption: $<$\textit{sample caption}$>$ 
\end{minipage}
}
                   \\
\bottomrule
\end{tabular}
\end{table}

%%%%%%%%%%%%%%%%%%%%%%%%%%%%%%%%%%%%%%%%%%%%%%%%%%%%%%%%%%%%%%%%%%%%%%%%%%%%%%%%%%%%%%%%%%%%%%%%%%%%%%%%%%%%%%%%%%%%%%%%
\end{document}